\documentclass[conf]{new-aiaa}
\usepackage[utf8]{inputenc}

\usepackage{graphicx}
\usepackage{amsmath}
\usepackage[version=4]{mhchem}
\usepackage{siunitx}
\usepackage{longtable,tabularx}
\usepackage{float}
\usepackage{qtree}
\usepackage{svg}
\usepackage{subcaption}
\usepackage{multirow}
\usepackage{booktabs}

\title{DreamLander: A 6-DOF World Model based Reinforcement Learning Planetary Lander}

\author{Nehma, G.,\footnote{PhD Student, Department of Aerospace, Physics and Space Sciences, gnehma2020@my.fit.edu}, Quinn, P.,\footnote{PhD Student, Department of Aerospace, Physics and Space Sciences, pquinn2019@my.fit.edu}, Tiwari, M.\footnote{Assistant Professor, Department of Aerospace, Physics and Space Sciences, mtiwari@fit.edu} and Riano-Rios, C.\footnote{Assistant Professor, Department of Aerospace, Physics and Space Sciences, crianorios@fit.edu}}
\affil{Florida Institute of Technology, Melbourne, Florida, 32901}

\begin{document}

\maketitle

\begin{abstract}

Soft powered descent landings are becoming evermore important in the field of space exploration and scientific advancement, because of their ability to deliver mission critical payloads, both highly accurately and safely. The high efficiency restriction on planetary landings due to the severely limited fuel supply results in the need for optimal guidance trajectories and control. Model based reinforcement learning is proving to be an important tool in simplifying the highly complex process of generating optimal guidance trajectories and control. In this work, we present DreamLander, a state of the art model based reinforcement learning algorithm based on DreamerV3, to generate optimal guidance and control for a realistic 6 degree of freedom entry descent landing problem. The model based approach creates a sample efficient and highly generalizable model that can be applied to a variety of scenarios whilst maintaining a high degree of accuracy, landing within 2 meters of the designated landing zone, and with velocity norm less than 0.3 meters per second. DreamLander makes use of state sensor observations, making progress toward End-to-End solutions of the Entry Descent Landing problem.
 
\end{abstract}


\section{Introduction}

Landing on extraterrestrial bodies, be it the Moon, other planets or even asteroids is a crucial part of any exploratory mission, and the task must be achieved safely and efficiently. Being able to safely deliver, resources, astronauts or robots to the surface of a celestial body is becoming even more paramount in missions, as seen with NASA's Discovery and Artemis Programs, JAXA's SLIM lander, SpaceX's recapture of launch vehicles, as well as other upcoming missions, and with that the need for highly accurate, automated, reproducible guidance and control systems are becoming evermore sought after. Because of the delicate nature of the equipment and payloads onboard the lander, hard landing methods such as the use of parachutes and inflatable balloons are often discarded for more favorable soft, powered entry descent landings (EDL's). A major benefit of such powered landings is that they can be autonomously guided and controlled \cite{D-Souza1997-fh, Bai2020-lo} such that a high degree of accuracy (within 5 meters of the target) can be achieved as well as small approach velocities of around 2 meters per second.

Automated guidance methods for the powered descent (PD) phase of the landings have been investigated since the Apollo program \cite{Meditch1964-jo, Klumpp1974-vn, apollo}, however due to the severely limited computational capability of that era, complex constraints on the optimization problems resulted in the use of polynomial guidance methods rather than optimal control theory. As computational prowess increased and algorithms in nonlinear programming \cite{10.5555/265009, Boyd_Vandenberghe_2004, Bomze2010-re} improved the capability of these guidance and control algorithms by allowing the deviation from explicit guidance laws, simplified models, or process constraints \cite{Biegler2009-rk, Domahidi2013-hp, Reynolds2017-gu}. Further improvements introduced variational gravitational acceleration and aerodynamic forces to the EDL problem, resulting in methods such as successive convex programming \cite{Lu2013-ah, Liu2014-kl}, the Jet Propulsion Laboratory (JPL) Guidance for Fuel-Optimal Large Diverts (G-FOLD) \cite{Acikmese2013-be, Scharf2014-zy, Scharf2017-qz}, Legendre pseudospectral method \cite{Sostaric2005-js} which all aim to improve the performance of the optimal guidance through convexification of the non convex problem. Control laws such as model predictive control (MPC), receding horizon control (RHC) and  zero-effort-miss/zero-effort-velocity may then be used in conjunction with the optimal guidance algorithms to efficiently and accurately maneuver the lander vehicle \cite{Sibley2010-re, D-Souza1997-fh, Song2020-no}.

With the recent popularity of neural network based approaches in a wide range of scientific fields, due to their powerful predictive and modeling capabilities, \cite{Tiwari2023-gu, Nehma2025-tf, Quinn2025-hg} it is no surprise that deep reinforcement learning (RL) algorithms have also been used in the discovery of guidance and control algorithms for the EDL problem \cite{Xu2021-rt, Ciabatti2021-tj}. Deep RL is a class of machine learning control optimization algorithms that can be broken down into either model-based or model-free, each with its own advantages and disadvantages. Unlike Imitation Learning (IL) where an expert trajectory is used to guide the learning of the IL agent, RL algorithms make use of unsupervised learning in order to arrive at an optimal policy that maximizes the given reward function for the problem \cite{Fujimoto2018-sg, Mnih2013-ps, Haarnoja2018-pi}. This results in a policy that is more robust and has the potential to learn and generalize to new environments far more quickly that an IL agent. Subclasses of RL include model-free algorithms such as REINFORCE \cite{Williams1992-mq}, PPO \cite{Schulman2017-mq}, DDPG \cite{Silver2014-uu}, TRPO \cite{Schulman2015-wg} etc. As the name suggests, these classes of models do not learn a representative model of the dynamics of the world but simply learn to output actions based on the observations presented. The benefits of such a model is that is often easier to implement and can perform better in more complex environments \cite{Ha2018-jl, Hafner2023-fy}, however it is usually less sample efficient than model based RL algorithms. Model based RL however, is usually more sample efficient as well as allows the agent to plan ahead using certain outcomes and can generalize to different contexts and tasks within environments \cite{Rimon2024-iv, chen2023selfsupervised}, as well as include methods  for safety and stability assurances\cite{10.1609/aaai.v33i01.33013387,8794107,NIPS2017_766ebcd5}. 

A number of methods in developing high fidelity planetary lander agents using model free RL have been developed previously, using algorithms such as DDPG \cite{Ciabatti2021-tj}, TD3 and SAC \cite{Xu2021-rt} and PPO \cite{Gaudet2020-ok} and show success both in generalization and accuracy, however the use of a model-based approach is largely understudied, with only an implementation of the DreamerV3 model being applied within the Space Robotics Bench benchmark \cite{orsula2025spaceroboticsbenchrobot}. \cite{Ciabatti2021-tj} employs transfer learning to generalize from the trained Martian environment to an astroid environment, but the highly generalizable properties of DreamerV3 suggest that with little hyperparameter tuning or few shot learning, our agent can successfully land on multiple environments. Our contributions in this work are as follows:

\begin{itemize}
    \item We propose a DreamerV3 based MBRL model to solve the late stage EDL problem using realistic physical parameters from the Apollo Moon landings. 
    \item Our model, DreamLander, is able to successfully solve the problem with higher performance and stricter accuracy than the original missions, ensuring greater safety. 
    \item We integrate realistic sensor based observations in the solution, making significant progress toward the full End-to-End approach to realistic, space robotics based problems. 
\end{itemize}

\section{Methodology}
 In this paper we develop DreamLander, which utilizes the popular and successful DreamerV3 model based reinforcement learning (MBRL) algorithm in order to generate optimal guidance trajectories and a control policy for EDL problems. DreamLander can be used in an End-to-End fashion to autonomously guide and control the lander to a successful, highly precise landing or, the generated guidance trajectories can be easily tracked by simple, reliable controllers such as a PID, adaptive, or MPC. The benefit of using the DreamerV3 algorithm is that, unlike model-free algorithms, because DreamerV3 learns a representation of the world through its latent vector, it is possible to generalize to new and varying environments with little hyperparameter tuning (few or zero shot adaptation \cite{xiong2025mrcommetaregularizedworldmodelgeneralizing, rimon2024mambaeffectiveworldmodel}). The robustness of this particular world model is that it can quickly be transferred to varying terrains, bodies and external forces so that a generalized framework that is suited for any range of EDL problems can be learned. We also note that due to the existence of the world model and latent space representation of the dynamics, that more rigorous and provable safety and stability guarantees with probabilistic guarantees can be formulated. The ability to ensure safety of the entire system can be performed in numerous methods including, Control Barrier Functions (CBFs) \cite{Cohen_2023, mahmud2021safety}, Hamiliton-Jacobi Reachability Analysis \cite{nakamura2025generalizing,8794107} or Lagrangian method constrained Markov Decision Processes \cite{lindner2024learningsafetyconstraintsdemonstrations, Hsu__2021}.
 
 In the original work \cite{Hafner2023-fy}, DreamerV3 was shown to generalize and work well for a range of different environments, ranging from state based locomotion to pixel based 2D games, each with a number of tasks. However, many of the environments that DreamerV3 was applied to, were either toy problems or simple demonstrative physical problems \cite{Hafner2023-fy}. The application of this framework to challenging, complex engineering problems has only recently begun to be explored,  seen in the case of autonomous driving scenarios \cite{Gao2024-zx, Zhao2024-cd}, but as of yet very limited applications into safety critical systems such as spacecraft and planetary landers have been conducted \cite{orsula2025spaceroboticsbenchrobot}. We aim to show that world model RL frameworks such as DreamerV3 are usable as guidance and/or control systems in these space-based applications.

In a conventional guidance, navigation and control (GNC) pipeline, a number of systems must be developed in unison to complete the full stack on-board computer, such as state estimation, sensor fusion, measurement filtering and the controller development, which is both highly complex and time consuming. With the advent of RL models, a new approach to the GNC pipeline, known as End-to-End, in which the RL agent receives raw observations such as camera imagery, LiDAR data, IMU and other sensors, and outputs directly the next required control, can be designed. It has been shown that DreamerV3 can be utilized in this End-to-End fashion, and given we can solve the optimal guidance problem with the states of the system, hence the next step would be to integrate DreamerV3 in an End-to-End fashion. This objective can be supported through NVIDIA's IsaacLab, which is built on Isaac Sim, a simulation framework with high fidelity simulation, rendering, asset management and compatibility with RL algorithms. 

In this work, we partially integrate raw sensor data into the observations, complimented with state data, to achieve the desired performance. We utilize the built-in Ray-Cast sensor which operates similar to a LiDAR to achieve the height mapping of the lander over the terrain. We incorporate an IMU to deliver velocity components, as well as a contact sensor on the legs of the lander to detect whether contact with the surface has been made and for how long contact has been made. Each of the sensors contain realistic noise levels with a mean noise of $0.08\mathrm{m}$ for the LiDAR and $0.01\mathrm{ms^{-1}}$ and $0.04^\circ s^{-1}$ for velocity and angular velocity respectively on the IMU.

\subsection{Simulation Framework}
\begin{figure}[h]
    \centering
    \includegraphics[width=0.7\textwidth]{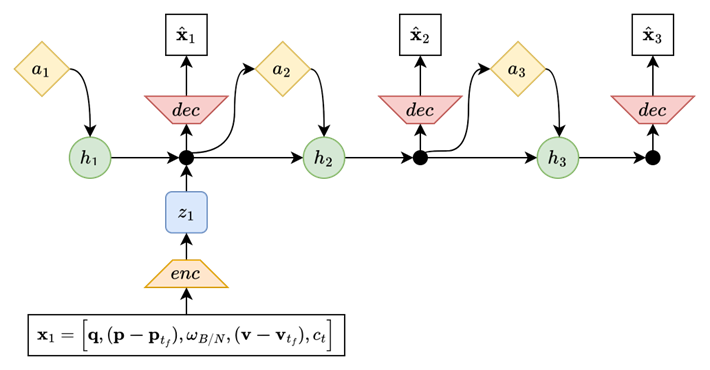}
    \caption{Process diagram showing the latent imagination of the model from an initial state to further states and the input of only the latent states to the agent to determine the action.}
    \label{fig:process}
\end{figure}

 DreamLander is built on the Dreamer world model framework, first popularized as a method of RL in \cite{Ha2018-jl}. It consists of a recurrent neural network (RNN) that is able to learn and predict a representation of the environment in a specified latent space. It then uses this latent representation of the world to perform two tasks, the first being passing this latent information of the observations to the actor-critic policy that is responsible for determining the next best action to take to achieve the goal. And second, because of the inherent memory of the RNN, the world model can be used to create new latent states of the future, similar to the process of dreaming, and then use these imagined latent states to further train the actor-critic policy. A representation of the process for both tasks is given in Figure \ref{fig:process}.

 The environment upon which DreamLander trains on is based on the 6-degrees-of-freedom (6-DOF) equations of motion between the lander and the desired landing point, which are elaborated on in the next section. Because we use IsaacLab for the construction of the environment in which we train DreamLander, we can make use of the NVIDIA PhysX engine that is built-in to IsaacLab to handle the dynamics propagation. PhysX integrates collision capabilities and full 6-DOF constraint support, which makes it an ideal candidate for our simulation with multiple sensors. We also focus our simulation on the near-landing phase of the EDL problem, hence we assume that the gravity vector points directly down. 

 The terrain generation within the simulation environment in IsaacLab is achieved using FBX files that are specifically designed and textured to have similar color and geometric features such as craters, mountains and rocks, to that of the desired planet or asteroid being simulated. Having high fidelity renders of the terrain are especially important for photorealistic observations from cameras, and help to reduce the Sim2Real gap \cite{verraest2025skydreamerinterpretableendtoendvisionbased}. In this work we do not include pixel based observations, so the quality of the rendering of the textures need not but ultra-high, but high fidelity geometric features are required, to enable accurate geometric representation of the terrain. 

\begin{figure}[H]
    \centering
    \begin{subfigure}[t]{0.4\textwidth}
        \centering
        \includegraphics[width=\linewidth,height=4.1cm]{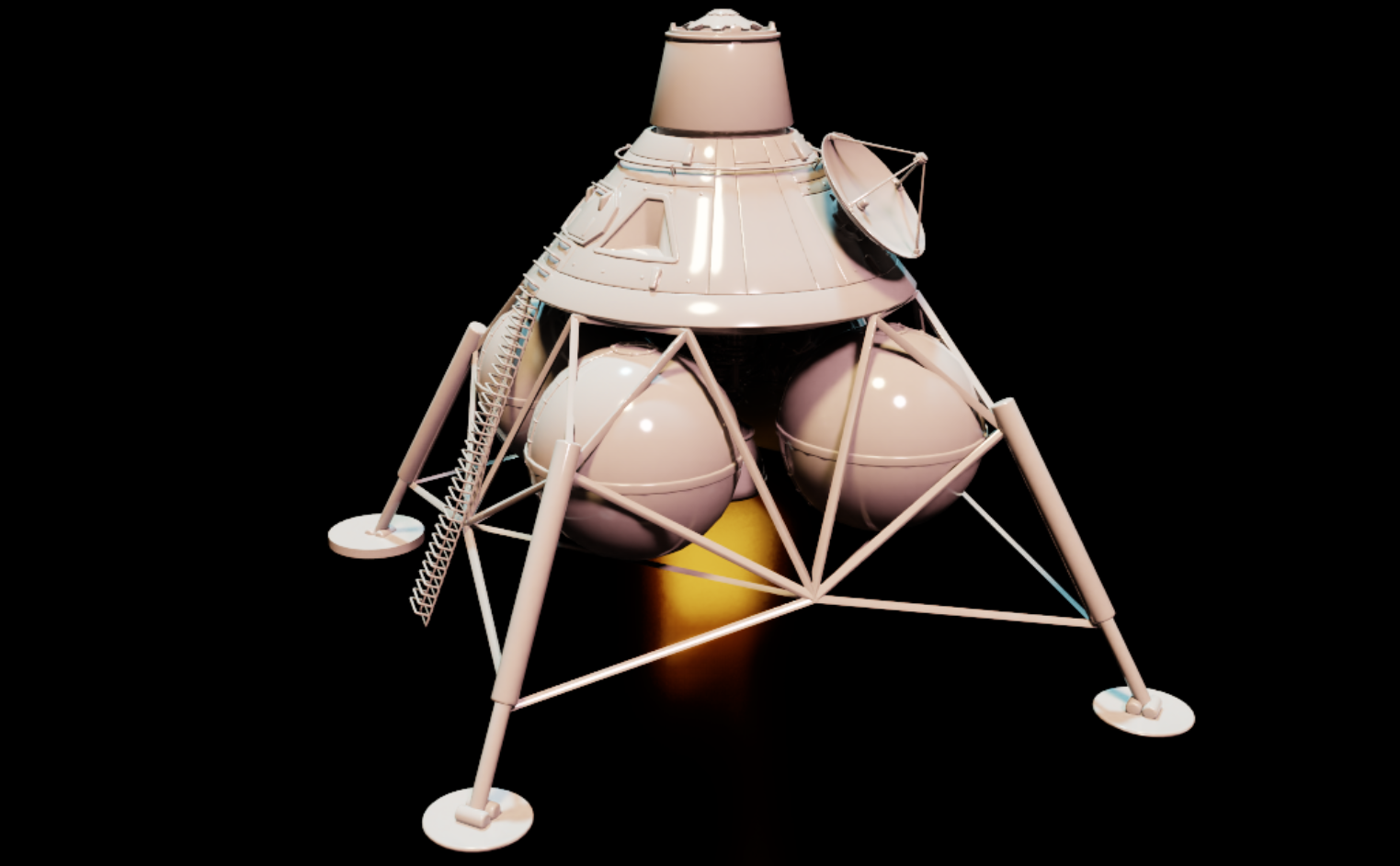}
        \caption{3D model of the lander used in simulation. The flame is purely for visual effect.}
       \label{fig:lander}
    \end{subfigure}
    ~
    \begin{subfigure}[t]{0.45\textwidth}
        \centering
        \includegraphics[width=\linewidth]{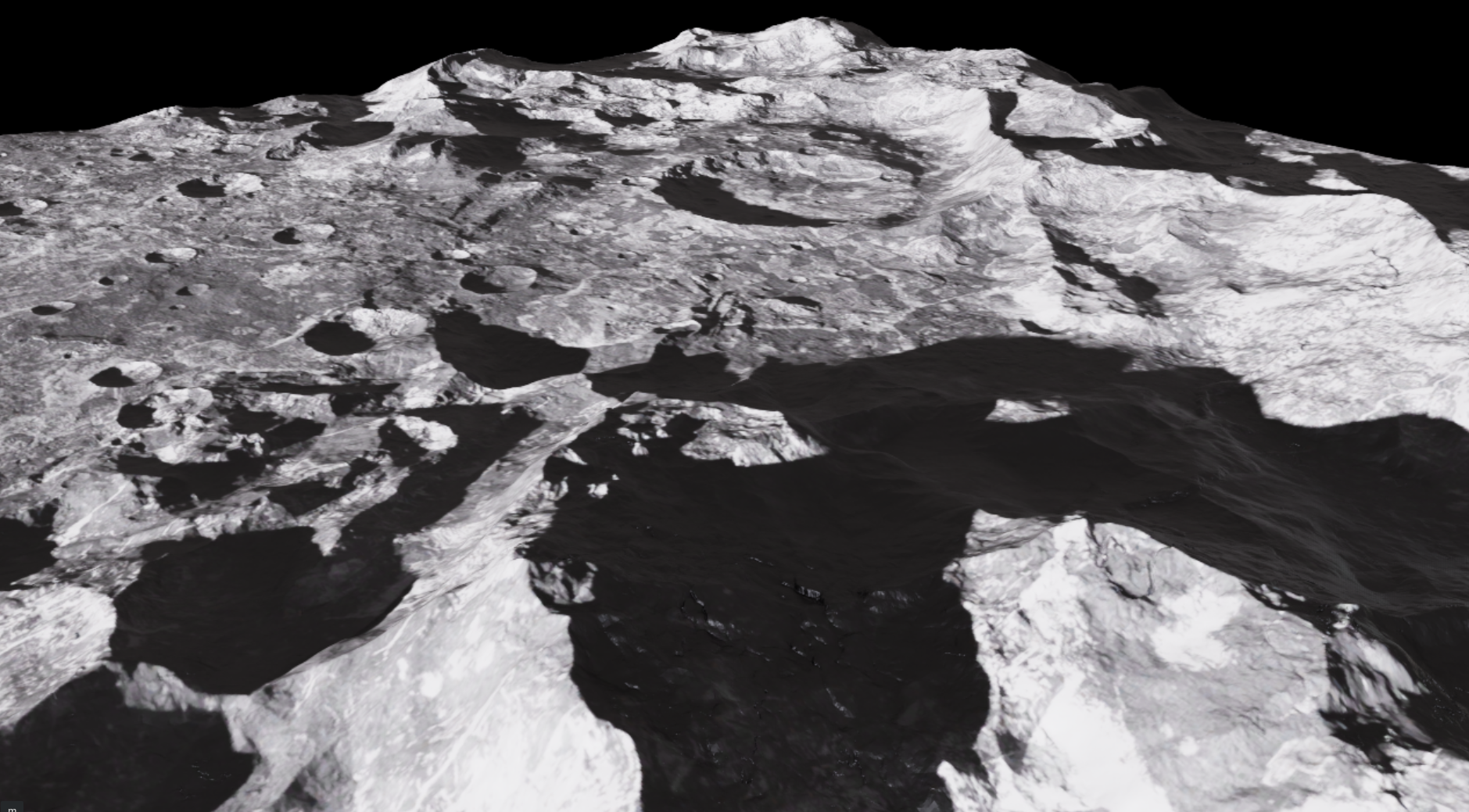}
        \caption{3D model of the terrain used in simulation, emphasizing the shadows, craters, mountains and valleys that improve realism of the simulation.}
       \label{fig:terrain}
    \end{subfigure}
    \caption{3D models of the lander and terrain used to simulate the environment of an EDL problem in IsaacLab.}
    \label{fig:3dmodels}
 \end{figure}

 The 3D model of the lander is another important aspect of the simulation framework. IsaacLab internally computes the inertia tensor of the vehicle based on user-defined mass or density parameters, so the choice of a lander that has geometrically similar features to that of a real vehicle, improves the overall fidelity of the simulation. We do not include a texture on the lander vehicle model as this does not impact the simulation. The Moon terrain and lander model can be visualized in Figures \ref{fig:lander} and \ref{fig:terrain}.

\subsection{Entry Descent Landing Dynamics}

Because of the fact that for many powered entry descent landing missions, a heat shield is utilized to protect the vehicle from the harsh environment of the initial reentry phase, and this shield is often jettisoned close to the ground with a short amount of time until landing, the dynamics of this problem are often only considered shortly after the jettison of the heat shield. As such, this results in the need for a control system that can orient itself quickly and provide robust control authority and guidance to make a safe and highly precise landing.

\begin{figure}[h]
    \centering
    \hspace{1cm}
    \includegraphics[width=0.4\linewidth]{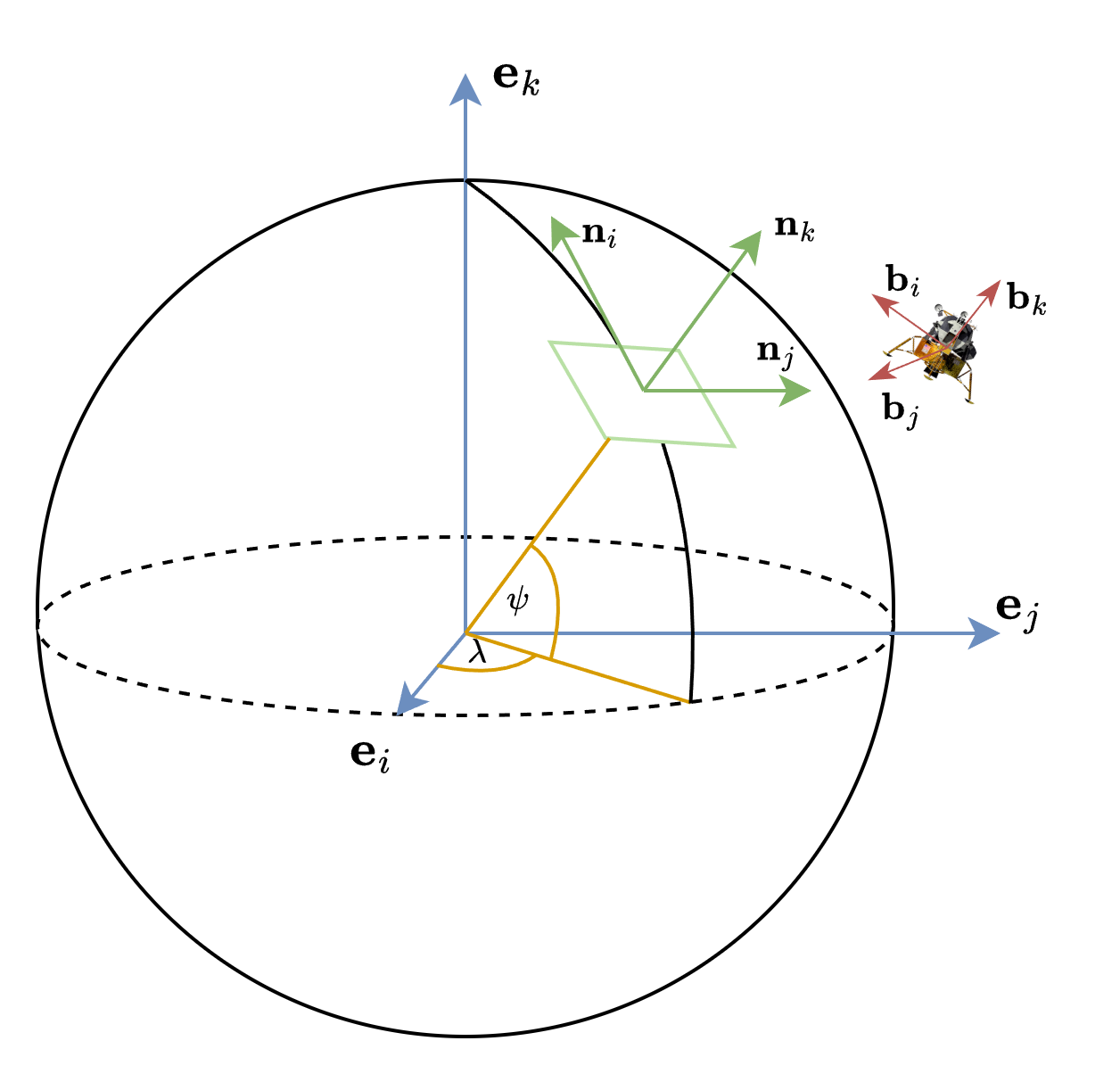}
    \caption{Coordinate frames for the 6-DOF environment used to represent the dynamics of the EDL problem.}
    \label{fig:frames}
\end{figure}

As the vehicle is near to the surface of the body it is landing on, we can make use of three reference frames to provide a complete description of the problem. The first is the fixed inertial frame with respect to the celestial body, similar to the Earth Centered Earth Fixed (ECEF) frame on Earth. The origin of the frame, $\mathcal{E}$ is at the center of the body, $\mathbf{e}_k$ is aligned with the axis of rotation of the celestial body, $\mathbf{e}_i$ points East along the equator of the body and $\mathbf{e}_j$ completes the right handed coordinate system. We can the define a second reference frame, the  East North Up (ENU) frame, which has its origin at the location of the landing zone (LZ), and is oriented by rotating the Inertial frame by the desired Longitude $\phi$ and Latitude $\lambda$. For the purposes of the EDL simulation, the ENU frame is considered to be inertially fixed and denoted by $\mathcal{N}$ with unit vectors $\mathbf{n}_i, \mathbf{n}_j, \mathbf{n}_k$. Finally the body frame is defined with its origin $\mathcal{B}$ at the center of mass of the vehicle, the z-axis $\mathbf{b}_k$ pointing upward out of the vehicle, the x-axis $\mathbf{b}_i$ pointing toward the front of the vehicle and the y-axis $\mathbf{b}_j$ completing the right hand coordinate system. The three coordinate systems can be visualized in Figure \ref{fig:frames}.

The lander itself has a number of key physical properties that are assumed based on historical information from the Apollo missions \cite{apollo} and prior works \cite{Cox-ab}. The mass $m$ of the vehicle is a constant $15000$ kg whilst the vehicle encompasses a $4\mathrm{m}\times4\mathrm{m}\times4\mathrm{m}$ volume and the lander's inertia tensor calculated by IsaacLab is given as:

\begin{equation} \label{eqn:inertiatensor}
    \mathbf{J} = \begin{bmatrix}
        18743.494 & -142.729 & 388.875 \\
        -142.729 & 18141.168 & 439.094 \\
        388.875 & 439.094 & 20367.090
    \end{bmatrix} kgm^2
\end{equation}

\begin{figure}[h]
    \centering
    \includegraphics[width=0.5\linewidth]{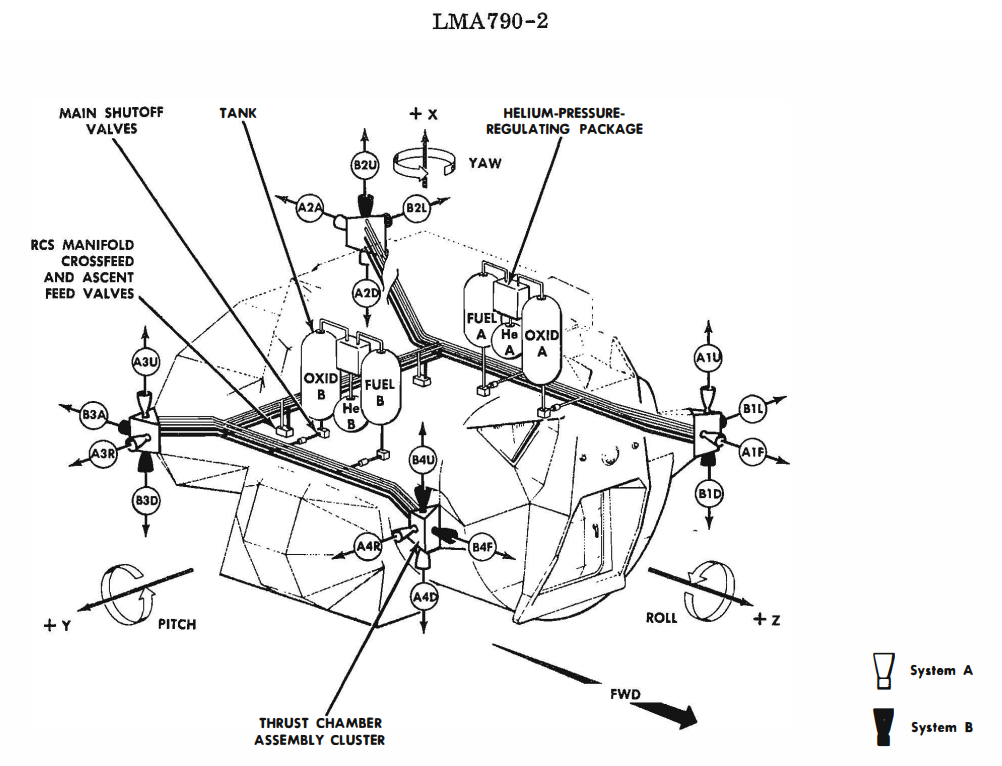}
    \caption{Diagram of the Reaction Control Subsystem thruster configuration of the Lunar Module Descent Stage \cite{vfm}.}
    \label{fig:RCSthrusters}
\end{figure}

 Even though it is assumed that the lander expels propellant to generate thrust, the mass and inertia tensors are kept constant in this simulation. The thruster arrangement on our model is assumed to be similar to that of the Apollo lander, with one main thruster beneath the lander and 4 small RCS thruster clusters at each corner of the lander as shown in Figure \ref{fig:RCSthrusters}. DreamLander is designed to output actions that correspond to the total forces and moments in the body frame as a result of all thrusters, since the alignment of the thrusters with respect to the body frame is constant at all times. Given that each RCS thruster is capable of a thrust range of 0N to 400N, the Main Engine is throttle-able between 0N to 43000N \cite{vfm} and the mounting configuration, it can be calculated that the range of forces and moments in the body frames are as shown in Table \ref{tab:minmaxinputs}.

 \begin{table}[h]
     \centering
     \begin{tabular}{c|c|c}
          Input & Min & Max \\
          \hline
          \hline
          $F_x$ ($N$) & -800 & 800 \\
          $F_y$ ($N$) & -800 & 800 \\
          $F_y$ ($N$) & 0 & 43000 \\
          $M_x$ ($Nm$) & -3200 & 3200 \\
          $M_y$ ($Nm$) & -3200 & 3200 \\
          $M_z$ ($Nm$) & -3200 & 3200 \\
     \end{tabular}
     \caption{Minimum and Maximum values of force and moment that the lander vehicle is capable of in the body frame.}
     \label{tab:minmaxinputs}
 \end{table}

The relationship between the force that is produced by each engine and the total force and moment that the vehicle experiences is given by, 

\begin{equation}
    \mathbf{F}^B = \sum^n_{i=1}\mathbf{d}_iT_{cmd,i}
\end{equation}

\begin{equation}
    \mathbf{M}^B = \sum^n_{i=1}\mathbf{r}_i\times\mathbf{F}^B_i
\end{equation}

Where, $T_{cmd}$ is the thrust commanded by each of the 16 RCS thrusters and main engine, $(\cdot)^\mathbf{B}$ and $(\cdot)^\mathbf{N}$ denote whether the vector is in the body or inertial frame and $\mathbf{d}_i$ and $\mathbf{r}_i$ are the unit thrust direction vectors and position vectors, in the body frame, of each thruster respectively. The definitions for both $\mathbf{d}_i$ and $\mathbf{r}_i$ are given in Tables \ref{tab:Thruster directions} and \ref{tab:Thruster positions}.

\begin{table}[!h]
    \centering
    \begin{minipage}[t]{.45\linewidth}
        \centering
        \begin{tabular}{c|c}
          Thruster & Unit Direction Vector $\mathbf{d}_i$ \\
          \hline 
          Main Engine & $\left[0 \quad 0 \quad 1 \right]$ \\
          RCS FR1 & $\left[0 \quad 0 \: -1 \right]$ \\
          RCS FR2 & $\left[0 \quad 1 \quad 0 \right]$ \\
          RCS FR3 & $\left[0 \quad 0 \quad 1 \right]$ \\
          RCS FR4 & $\left[-1 \quad 0 \quad 0 \right]$ \\
          RCS FL1 & $\left[0 \quad 0 \: -1 \right]$ \\
          RCS FL2 & $\left[-1 \quad 0 \quad 0 \right]$ \\
          RCS FL3 & $\left[0 \quad 0 \quad 1 \right]$ \\
          RCS FL4 & $\left[0 \: -1 \quad 0 \right]$ \\
          RCS BL1 & $\left[0 \quad 0 \: -1 \right]$ \\
          RCS BL2 & $\left[0 \: -1 \quad 0 \right]$ \\
          RCS BL3 & $\left[0 \quad 0 \quad 1 \right]$ \\
          RCS BL4 & $\left[0 \quad 1 \quad 0 \right]$ \\
          RCS BR1 & $\left[0 \quad 0 \: -1 \right]$ \\
          RCS BR2 & $\left[1 \quad 0 \quad 0 \right]$ \\
          RCS BR3 & $\left[0 \quad 0 \quad 1 \right]$ \\
          RCS BR4 & $\left[0 \quad 1 \quad 0 \right]$ \\
        \end{tabular}
        \caption{Unit Direction vectors for each thruster. FR, FL, BL, BR designate front right, front left, back left, back right. RCS cluster is counted from top going clockwise looking out from inside the lander.}
        \label{tab:Thruster directions}
    \end{minipage}
    \hspace{1cm}
    \begin{minipage}[t]{.45\linewidth}
        \centering
        \begin{tabular}{c|c}
          Thruster & Position Vector $\mathbf{r}_i$ \\
          \hline 
          Main Engine & $\left[0 \quad 0 \: -1.4 \right]$ \\
          RCS FR1 & $\left[2 \: -2 \quad 0 \right]$ \\
          RCS FR2 & $\left[2 \: -2 \quad 0 \right]$ \\
          RCS FR3 & $\left[2 \: -2 \quad 0 \right]$ \\
          RCS FR4 & $\left[2 \: -2 \quad 0 \right]$ \\
          RCS FL1 & $\left[2 \quad 2 \quad 0 \right]$ \\
          RCS FL2 & $\left[2 \quad 2 \quad 0 \right]$ \\
          RCS FL3 & $\left[2 \quad 2 \quad 0 \right]$ \\
          RCS FL4 & $\left[2 \quad 2 \quad 0 \right]$ \\
          RCS BL1 & $\left[-2 \quad 2 \quad 0 \right]$ \\
          RCS BL2 & $\left[-2 \quad 2 \quad 0 \right]$ \\
          RCS BL3 & $\left[-2 \quad 2 \quad 0 \right]$ \\
          RCS BL4 & $\left[-2 \quad 2 \quad 0 \right]$ \\
          RCS BR1 & $\left[-2 \: -2 \quad 0 \right]$ \\
          RCS BR2 & $\left[-2 \: -2 \quad 0 \right]$ \\
          RCS BR3 & $\left[-2 \: -2 \quad 0 \right]$ \\
          RCS BR4 & $\left[-2 \: -2 \quad 0 \right]$ \\
        \end{tabular}
        \caption{Position vectors for each thrusters from the center of mass of the vehicle.}
        \label{tab:Thruster positions}
    \end{minipage}
\end{table}

 As DreamLander only outputs total force and moment control inputs, it is assumed that the total force or moment required is evenly split between each of the thrusters responsible for producing that force or moment. For example, if a desired yaw moment of $1600\mathrm{Nm}$ around the body z-axis was commanded, then RCS thrusters FR2, FL2, BL2, BR2 would each contribute $200\mathrm{N}$ of thrust to achieve the desired yawing moment.

The equations of motion for the lander, upon approach to the LZ can now be described by the traditional 6-DOF equations of motion using quaternions. Although it is unlikely that the vehicle will reach a state where gimble-lock becomes an issue, during training of DreamLander, a large space of the state space is often searched due to initially high entropy values, hence it is imperative that the dynamics can account for this. It is important to note that the states referenced by the dynamics are not the same states that are referenced by the RL algorithm, as in, the states used by DreamLander are the observations from the environment. As such we denote the state space state vector as:

\begin{equation} \label{eqn:state_vec}
    \mathbf{x}=\begin{bmatrix}
        \mathbf{q}^T & \mathbf{p}^T & \mathbf{\omega}_{B/N}^T & \mathbf{v}^T
    \end{bmatrix}^T
\end{equation}

Where $\mathbf{q}$ is the quaternion that represents the rotation between the body frame $\mathcal{B}$ and the inertial ENU coordinate frame $\mathcal{N}$, $\mathbf{p}$ is the position vector of the lander in the inertial frame, $\mathbf{\omega}_{B/N}$ is the angular velocity of the lander in the body frame, and $\mathbf{v}$ is the velocity of the lander in the inertial frame. The control input can then be defined as

\begin{equation}
    \mathbf{u}=\begin{bmatrix}
        \mathbf{F^B}^T & \mathbf{M^B}^T
    \end{bmatrix}^T.
\end{equation}

Hence, the controlled, nonlinear, continuous-time dynamical system is propagated according to the following mathematical expression

\begin{equation}
    \label{eqn:dynamics}
    \dot{\mathbf{x}}=f(\mathbf{x},\mathbf{u}).
\end{equation}

To evolve the dynamics of the lander, the PhysX engine uses a semi-implicit Euler Integration method, with a Temporal Gauss-Seidel (TGS) solver in order to account for the contact dynamics that are not included in Equation \ref{eqn:dynamics} \cite{isaaclabdocs}. Hence, the time derivatives of each of the states are required and can be given as follows:

\begin{equation}
    \dot{\mathbf{q}}=\frac{1}{2}\mathbf{q}\ast\mathbf{\omega}_{B/N}
\end{equation}

\begin{equation}
    \dot{\mathbf{p}}=\mathbf{v}
\end{equation}

\begin{equation}
    \dot{\mathbf{\omega}}_{B/N} = \mathbf{J}^{-1}\left(\mathbf{M^B} - \mathbf{\omega}^\times_{B/N}\mathbf{J}\mathbf{\omega}_{B/N}\right)
\end{equation}

\begin{equation}
    \dot{\mathbf{v}} = \frac{\mathbf{F^N}}{m}+\mathbf{g}
\end{equation}

Where, $\ast$ represents the quaternion product. Since quaternions are of dimension 4 and vectors are of dimension 3, the vectors must have a 0 appended in front of them to allow for valid quaternion multiplication. $\mathbf{J}$ is the inertia tensor defined in Equation \ref{eqn:inertiatensor}, $\mathbf{g}=[0,0,-1.62]^T\mathrm{ms^{-2}}$ is the gravity vector and $m$ is the mass of the vehicle. 

The expression $\mathbf{a}^\times$ denotes a skew-symmetric matrix that is equivalent to performing the cross product and is defined as:

\begin{equation}
    \mathbf{a}^\times = \begin{bmatrix}
        0 & -a_3 & a_2 \\ a_3 & 0 & -a_1 \\ -a_2 & a_1 & 0
    \end{bmatrix}
\end{equation}

Finally, the value for $\mathbf{F^N}$ can be found via the following relation between the body frame force,  $\mathbf{F^B}$ and the quaternion representing the rotation between the body and inertial frame:

\begin{equation}
    \begin{bmatrix}
        0 \\ \mathbf{F^N}
    \end{bmatrix} = \mathbf{q} \ast \begin{bmatrix}
        0 \\ \mathbf{F^B} 
    \end{bmatrix} \ast \mathbf{q}^{-1}
\end{equation}

The simulation and control policy that results is run at 10Hz, whereas the solver and simulation run at 60Hz to ensure stable and smooth dynamics and solutions to the differential equations. 

\subsection{DreamerLander Architecture}

\begin{figure}[h]
    \centering
    \includegraphics[width=0.60\linewidth]{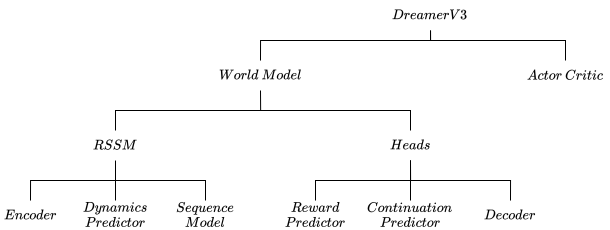}
    \caption{Architecture of the DreamerV3 WM.}
    \label{fig:dreamer}
\end{figure}

 DreamLander is built from DreamerV3 which is a world-model based RL algorithm that incorporates a autoencoder model to map observations to and from the latent space, a Recurrent State Space Model (RSSM) to learn and propagate the dynamics of the system in the latent space as well as a simple Actor-Critic agent to learn and execute the controls required to succeed in the given task. The high-level breakdown of the World Model (WM) includes the RSSM, which is comprised contains the encoder, dynamics predictor and sequence model, as well as a series of \textit{heads} which are the reward and continuation predictors and decoder shown in Figure \ref{fig:dreamer}

 The autoencoder needs to map raw observation inputs from the observation space to the usually lower dimensional latent space,  hence for simulations in which the observations arrive as images, the pixels are encoded and decoded through a convolutional neural network (CNN) autoencoder. If the observation space is a vector then the encoding and decoding is performed via a fully connected multilayer perceptron (MLP) network. 

 The structure of the DreamerV3 network is closely based on the architecture of the DreamerV2 \cite{Hafner2020-jm} world model that was shown to have good predictive accuracy on discrete world models. The RSSM \cite{Hafner2019-pt}, consists of multiple smaller networks that all play a specific role. As shown in Figure \ref{fig:process} there is a deterministic recurrent state $h$ that is an output of the RNN, which stores memory from past observations and states. At each step, the recurrent state $h_t$ is used to help predict a posterior latent state $z_t$ that contains information from the encoded image $x_t$, as well as a prior latent state $\hat{x}_t$ that does not include information from the current observation $x_t$ but is supposed to closely predict the posterior state $z_t$. In training this is enforced by including the KL divergence between $z_t$ and $\hat{z}_t$ in loss function. After training, since the prior state $\hat{z}_t$ can accurately predict what would be the latent state with the next image, this is what is used for dreaming trajectories, without actual observations. 

 The model then uses a concatenated state of both the deterministic RNN state $h_t$ and the stochastic posterior state $z_t$ to feed 3 separate networks that generate a prediction of the next observation $\hat{x}_t$, the reward $\hat{r}_t$ and the discount factor $\hat{\gamma}_t$. 

 The behavior policy of the RL agent is a simple actor critic framework that utilizes the REINFORCE \cite{Williams1992-mq} algorithm to iteratively improve the actor toward the optimal control policy. The actor and critic are trained solely on latent state representations of the world, generated by the world model, hence this allows for significantly faster training and the ability to learn in the imagined trajectories created by the learned world model. During training of the actor and critic, the world model is fixed. 

 The actor and critic are trained cooperatively, meaning that the actor generates actions that lead to states that maximize the value output of the critic, whilst the critic uses the actions and states generated by the actor to accurately estimate the sum of rewards. Hence they both have separate loss functions that they try to optimize. The critic is parameterized on $\psi$ whilst the actor is parameterized on $\theta$:

 \begin{equation}
     \mathcal{L}(\psi) = -\sum^T_{t=1} \text{ln}p_\psi(R^\lambda_t|s_t)
 \end{equation}

 \begin{equation}
     \mathcal{L}(\theta) = -\sum^T_{t=1} sg\left( (R^\lambda_t-v_\psi(s_t))/\text{max}(1,S)\right)\text{log}_{\pi_\theta}(a_t|s_t)+\eta H[\pi_\theta(a_t|s_t)]
 \end{equation}

Where,

\begin{subequations}
    \begin{equation}
        R^\lambda_t = r_t + \gamma c_t\left( (1-\lambda)v_t + \lambda R^\lambda_{t+1}\right)
    \end{equation}

    \begin{equation}
        R^\lambda_T=v_t
    \end{equation}

    \begin{equation}
        S = EMA\left(Per(R^\lambda_t,95)-Per(R^\lambda_t,5),0.99\right)
    \end{equation}
\end{subequations}

Here $R^\lambda_t$ is the distribution of the return estimates, $s_t$ is the imagined model states, $v_t$ is the critics predicted values, $a_t$ is the actor output, $\lambda$ and $\gamma$ are discount factors, $EMA$ is the exponential moving average, $Per(x,n^{th})$ is the $n^{th}$ percentile of some value $x$ and $H$ is the entropy of the model. Further details regarding the DreamerV3 architecture can be found in the original work \cite{Hafner2023-fy}.

\subsection{Reinforcement Learning Problem}

RL generates optimal solutions because of its nature of maximizing reward functions that have been specifically designed to improve the models behavior in reaching the goal. For any planetary lander, the objective function is to find the thrust program that minimizes the fuel consumption of the  based on the thrust $\mathbf{T}$, specific impulse $I_\mathrm{sp}$ and reference gravity $g_\mathrm{ref}$:

\begin{equation}
   \min_{\mathbf{T,t_f}} \mathcal{J} = \int^{t_f}_0 \frac{\left|\left|\mathbf{T}\right|\right|}{I_\mathrm{sp}g_\mathrm{ref}} dt
\end{equation}

Subject to the constraints from the dynamics of the system and boundary conditions:

\begin{subequations}
    \begin{equation}
        0 < T_{min} \le \left|\left|\mathbf{T}\right|\right| \le T_{max}
    \end{equation}
    
    \begin{equation}
        \mathbf{p}(0) = \mathbf{p}_0
    \end{equation}

    \begin{equation}
        \mathbf{v}(0) = \mathbf{v}_0
    \end{equation}

    \begin{equation}
        \mathbf{p}(t_f) = \mathbf{p}_{t_f}
    \end{equation}

    \begin{equation}
        \mathbf{V}(t_f) = \mathbf{v}_{t_f}
    \end{equation}
\end{subequations}

Where $\mathbf{p}$ and $\mathbf{v}$ are the position and velocity vectors of the lander in the inertial frame, respectively, $(\cdot)_0$ denotes the initial state and $(\cdot)_{t_f}$ denotes the final, desired state. In order to achieve this minimization of the objective function, the RL agent needs to have a reward function defined, whereby maximizing the reward function leads to a minimization of the objective function. For the EDL problem, constructing a reward function that is solely based on the optimization of the fuel leads to reward spaces that are both sparse and do not properly guide the agent to the right policy that optimizes the objective function. As such, the reward function that is used penalizes errors in the tracking position and velocity, penalizes excessive yaw, pitch and roll angles of the lander, penalizes excessive control effort, penalizes as the attitude approaches the attitude limits, rewards a bonus for successfully landing and penalizes a crash landing. Mathematically this looks like:

\begin{equation} \label{eqn:reward}
    \begin{split}
    r_t = \left(\mathbf{\psi}(t+\Delta t)-\mathbf{\psi}(t)\right) &+ \left(\frac{1}{\alpha_3}-\frac{1}{\alpha_3e^{(\frac{-\mathbf{\theta_e(t)}}{0.4})}}\right) + \alpha_4\left|\left| \mathbf{\omega}_{B/N}(t)\right|\right| + \alpha_5\left|\left| \mathbf{F^B}(t)\right|\right| + \alpha_6\left|\left| \mathbf{M^B}(t)\right|\right|\\ 
    &+ \alpha_7\text{if any}(hovering)  + \alpha_{landed} + \alpha_{landed+aligned} - \alpha_{crashed}
    \end{split}
\end{equation}

Where $\alpha_1-\alpha_7$ are weighting parameters for each of the rewards that are either positive or negative based on whether that portion of the reward function penalizes or rewards the agent and are detailed in Table \ref{tab:rewardparams}. $\alpha_{landed}$, $\alpha_{crashed}$, $\alpha_{landed+aligned}$ reward the agent for meeting the criteria of a successful landing, penalize for crashing and reward heavily for an aligned landing, respectively, $\psi(t)$ is the potential function used in the potential-based reward for both position and velocity, $\mathbf{e}_q$ is the error quaternion between the current and desired orientation, measuring the alignment of the vehicle with respect to its orientation and the conditions $\text{if any}(\mathbf{q}(t)\ge \mathbf{q}_{lim})$ and $\text{if any}(hovering)$ apply the respective $\alpha$ penalizes if those conditions are satisfied. $\alpha_6$ penalizes large deviations from alignment, whilst $\alpha_7$ penalizes the use of the thrusters when the vehicle is hovering above the LZ, to approximate shutting down the engines when close enough to land completely.

The use of potential-based reward shaping enables faster progress toward a goal without the notion of getting `distracted' \cite{Ng1999PolicyIU}. Thus, the potential function $\psi(t)$ is defined as 

\begin{equation}
    \psi(t) = \alpha_1\left|\left| \mathbf{p}-\mathbf{p}_{t_f}\right|\right|_D + \alpha_2\left|\left| \mathbf{v}-\mathbf{v}_{t_f}\right|\right|
\end{equation}

Where, to induce a better trajectory that discourages hovering at low altitude whilst adjusting the x-y position, the weighted norm $\lVert\cdot\rVert_D$ \cite{Olver2018} , is used to more strongly punish errors in the $x$ and $y$ directions. The intuition follows that because in this EDL problem, the distance from the LZ is significantly larger in the $z$ direction than the $x$ and $y$ directions, using an unweighted two-norm can potentially lead to an optimal policy in which the vehicle reduces its altitude faster than it can correct its error in the x-y directions. As such we define the weighted norm and its corresponding weighing matrix as

\begin{equation}
    \lVert\mathbf{p-p}_{t_f}\rVert_D=\sqrt{\langle(\mathbf{p-p}_{t_f});(\mathbf{p-p}_{t_f})\rangle}=\sqrt{(\mathbf{p-p}_{t_f})^TC(\mathbf{p-p}_{t_f})} \quad,\quad C=\begin{bmatrix}
        1 & 0 & 0 \\ 0 & 1 & 0 \\ 0 & 0 & 0.6
    \end{bmatrix}
\end{equation}

The second term in Equation \ref{eqn:reward} penalizes misalignment between the body axis and the reference axis at the LZ. For simplicity, and because we utilize the quaternion in the expression of the orientation, we define the reference quaternion as $\mathbf{q}_{des}=\begin{bmatrix}1&0&0&0\end{bmatrix}^T$. Hence the error quaternion can be found by 

\begin{equation}
    \mathbf{e}_q=\mathbf{q}_{des}\ast\mathbf{q}^*
\end{equation}

Where $\ast$ is the quaternion product and $\cdot^*$ denotes the quaternion conjugate \cite{6669617}. We then convert the error quaternion to axis-angle representation to find the angular error, $\mathbf{\theta}_e(t)$ in alignment between the two frames, as this can be used in the inverse exponential reward function to guide DreamLander to align the lander. The angle can be found via the following,

\begin{equation}
    \mathbf{\theta}_e(t)= 2atan2\left(\lVert\mathbf{e}_{q,v}\rVert,\left|e_{q,0}\right|\right)
\end{equation}

Where $\mathbf{e}_{q,v}$ and $e_{q,0}$ are the vector and scalar part of the error quaternion, respectively. If $\mathbf{\theta}_e(t)< 0.057$ then the alignment is within 3 degrees, and considered aligned.

Based on the reinforcement learning problem presented above, it is necessary to define an observation that can be computed based off the dynamical state, and fed into DreamLander. The observation is defined by the state vector given in Equation \ref{eqn:state_vec}, with the substitution of the position and velocity vectors with their corresponding error state,  concatenated with the contact sensor observation $c_t$, which indicates the current amount of time in contact between the feet of the lander and the terrain. Hence, the observation vector is then given as:

\begin{equation} \label{eqn:obs_vec}
    \mathrm{obs}=\begin{bmatrix}
        \mathbf{q} & (\mathbf{p-p}_{t_f}) & \mathbf{\omega}_{B/N} & (\mathbf{v-v}_{t_f}) & c_t
    \end{bmatrix}^T
\end{equation}

We again emphasize that both velocity (angular and linear) components of the state vector are obtained using the built-in IMU sensor within IsaacLab's simulation framework. The altitude, corresponding to the z-axis of the position vector $\mathbf{p}$, is obtained via a downward facing LiDAR through IsaacLab's Ray-Cast sensor, and the contact time is directly measured via the contact sensor that measures the intersection of boundaries between two objects in the simulation. The use of the error state in DreamLander's observation vector allows for dynamic selection of the LZ, which may prove useful with more complex guidance or pixel based guidance that DreamerV3 is capable of. 


\section{Simulation Results \& Discussion}

DreamLander is specifically trained on solving the EDL problem for end-stage landing. The training process is conducted through IsaacLab which enables the use of multiple environments to enhance and speed up data collection and training as well as the incorporation of multiple sensors that cannot be analytically expressed through dynamic equations. However, due to difference in how IsaacLab and the DreamerV3 architecture treat multiple environments (IsaacLab only allows the creation of one instance of the \textit{world}, with multiple environments inside that $\textit{world}$, whereas DreamerV3, which uses Gym environments, expects to be able to create separate instances of the environments), a wrapper to synchronize the two architectures is needed to ensure proper training that utilized the efficiency of all environments. We trained DreamLander on a single NVIDIA GeForce RTX 3090 GPU, with 8 environments and the training hyperparameters outlined in Table \ref{tab:hyperparams}. Training was completed on average in 17 hours. The related code can be found at: \url{https://github.com/george-nehma/IsaacLab-Dreamerv3}.

\begin{table}[h]
\centering
    \begin{minipage}[t]{.48\linewidth}
        \centering
         \begin{tabular}{c|c}
         \multicolumn{2}{c}{\textbf{Environment Parameters}} \\
         \hline
         \textbf{Parameter} & \textbf{Value} \\
         \hline
         $\alpha_1$   & -1.3\\
         $\alpha_2$ &   -1.3\\
         $\alpha_3$ & 10\\
         $\alpha_4$ & -0.05\\
         $\alpha_5$    &$\mathrm{diag}[-0.001\:-0.001\:-0.001]$\\
         $\alpha_6$    &$\mathrm{diag}[-0.3\:-0.3\:-0.3]$\\
         $\alpha_{landed}$ & 20\\
         $\alpha_{crashed}$ & -10\\
         $\alpha_{landed+aligned}$ & 50\\
         $\mathbf{p}_{land}$ & 2m\\
         $\mathbf{v}_{land}$ & 0.3$\mathrm{ms^{-1}}$\\
         $c_{land}$ & 0.5sec\\
         \end{tabular}
         \caption{Environment parameters for the Lunar Lander EDL problem}
         \label{tab:rewardparams}
        
    \end{minipage}
    \hspace{0.5cm}
    \begin{minipage}[t]{.48\linewidth}
        \centering
        \begin{tabular}{c|c}
             \multicolumn{2}{c}{\textbf{Training Hyperparameters}} \\
             \hline
             \textbf{Parameter} & \textbf{Value} \\
             \hline
             steps   & 5e5\\
             action\_repeat & 2\\
             prefill &   8000\\
             dyn\_hidden & 256\\
             dyn\_deter & 256\\
             units    & 1024\\
             batch\_size &18\\
             batch\_length & 128\\
             train\_ratio & 1024\\
             envs & 8\\
             \end{tabular}
             \caption{DreamLander main hyperparameters corresponding to the DreamerV3 PyTorch version \cite{Nm512}.}
             \label{tab:hyperparams}
    \end{minipage}
\end{table}

\subsection{Lunar Environment}

In all simulations, both training and evaluation, the initial conditions were randomly distributed from the following bounds: $x$ and $y$ position; $[-20\mathrm{m},20\mathrm{m}]$, $z$ position; $[130\mathrm{m},150\mathrm{m}]$, Euler angles $(\phi, \theta, \psi); \:[-10^\circ,10^\circ]$, $x$ and $y$ velocity; $[-2\mathrm{ms^{-1}},2\mathrm{ms^{-1}}]$, $z$ velocity; $[-5\mathrm{ms^{-1}},-1\mathrm{ms^{-1}}]$, angular velocities; $[-0.035\mathrm{rads^{-1}},0.035\mathrm{rads^{-1}}]$. Specifically for the Moon environment, the gravitational acceleration was set at a constant $-1.62\mathrm{ms^{-1}}$. In order to determine whether the vehicle successfully landed, limits on the terminal state of the simulation were imposed, with the LZ set as and the origin of the ENU frame, $\mathbf{p}_{t_f}=[0\: 0]^T$, the condition for landing is as follows

\begin{equation} \label{eqn:land_cond}
    \lVert\mathbf{p}_T\rVert\le\mathbf{p}_{land}, \quad \lVert\mathbf{v}_T\rVert\le\mathbf{v}_{land}, c_t\le c_{land}
\end{equation}

Where $\mathbf{p}_{land}$ is the limit on the distance from the desired landing point, and $\mathbf{v}_{land}$ is the limit on the landing velocities. It should be noted that these landing conditions are stricter than those of the Apollo missions \cite{apollo} and are deliberately chosen to emphasize the precise and high performance capability of MBRL algorithms. Each environment had a maximum time $T=90\,\mathrm{sec}$ however, once DreamLander wass trained and converged, all simulations were completed within $40\,\mathrm{sec}$. 

Our training session was run with a budget of 500,000 iteration steps as we found this to be long enough for DreamLander to learn the model of the world and converge to the optimal value for each of the individual losses. Within the DreamerV3 architecture, the $\textit{train ratio}$ hyperparameter controls the number of iteration steps taken per step of the actual environment. In our training, this value was set to 1024, with a batch size of 18 and batch length of 128. The RSSM dynamic predictor model has 256 GRU recurrent units, 1024 dense hidden units while the actor and critic heads each had 4 layers in their respective MLPs, with the continuation head only having 2 layers and the reward having 4. In total, this resulted in World Model network with 18,418,702 parameters, the actor with 4,476,940 and the critic with 4,726,015.

In order to evaluate our trained model, we twice ran 100 separate simulations, with randomly chosen initial conditions consistent with the training initial conditions, and landing criteria consistent with the conditions in Equation \ref{eqn:land_cond}. The first set of 100 trajectories had the LZ set to the origin of the ENU frame, whilst the second set had it's LZ set to the X-Y coordinates $[10,-5]$. Out of the 100 evaluation trajectories in the first set, 92 simulations successfully landed, whilst the other 8 simulations crashed. This resulted in DreamLander achieving a success rate of 98\%. It should be noted however, that of the 8 crashed trajectories, 2 did not meet the landing criteria solely because of the norm of landing velocity being between 0.14 and $0.98\,\mathrm{ms^{-1}}$ larger than the target landing velocity of $0.3\,\mathrm{ms^{-1}}$, 1 was unable to land within the precise landing radius despite having good velocities, and 5 missed both the landing velocity and accuracy criteria. For the two trajectories that missed the landing velocity, the velocity in the $z$ direction was satisfactory.

\begin{figure}[H]
    \centering
    \begin{subfigure}[t]{0.45\textwidth}
        \centering
        \includegraphics[width=\linewidth]{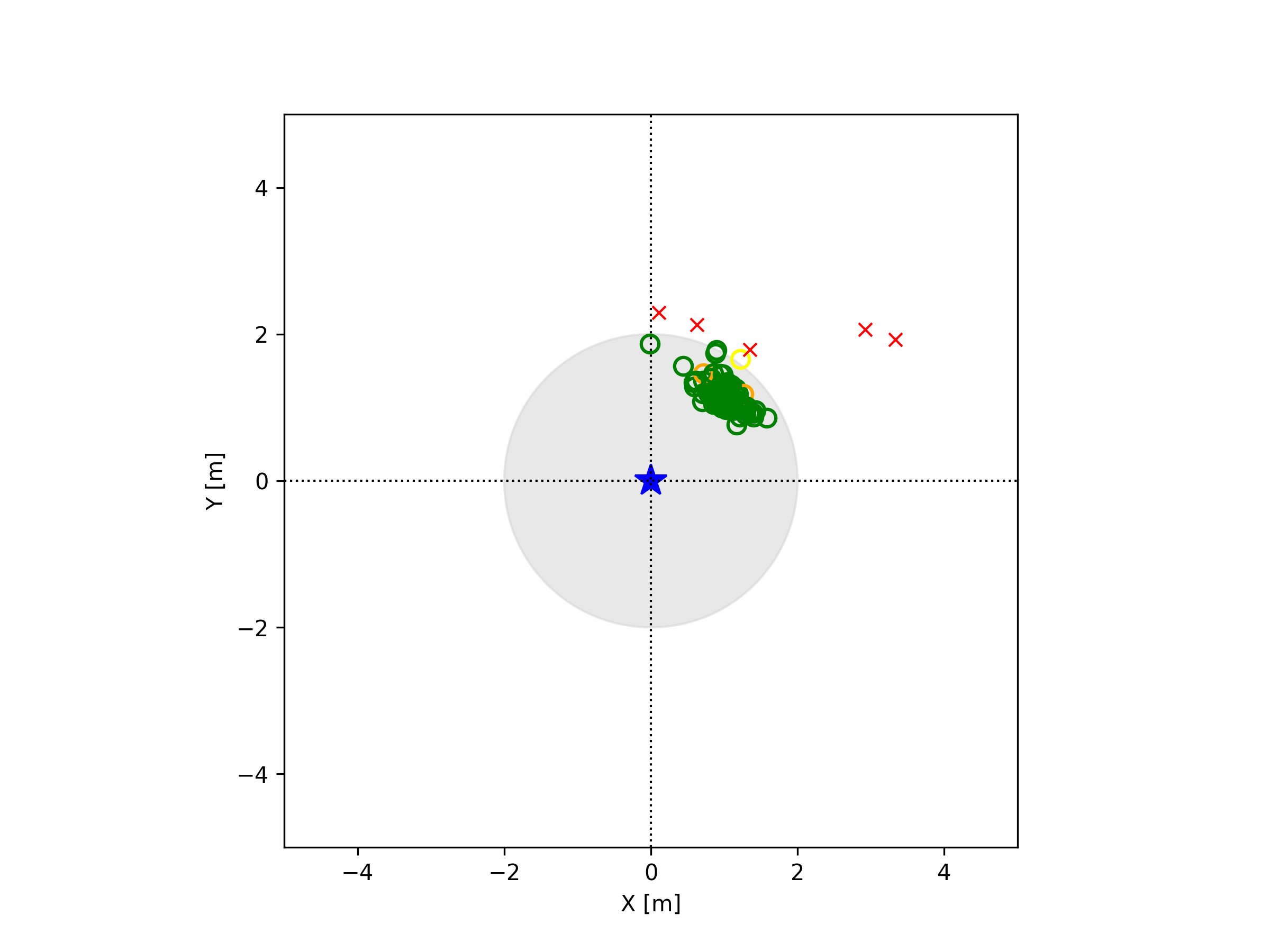}
        \caption{Landing points for the 100 simulations with LZ at the origin.}
       \label{fig:LZ_0}
    \end{subfigure}
    ~
    \begin{subfigure}[t]{0.45\textwidth}
        \centering
        \includegraphics[width=\linewidth]{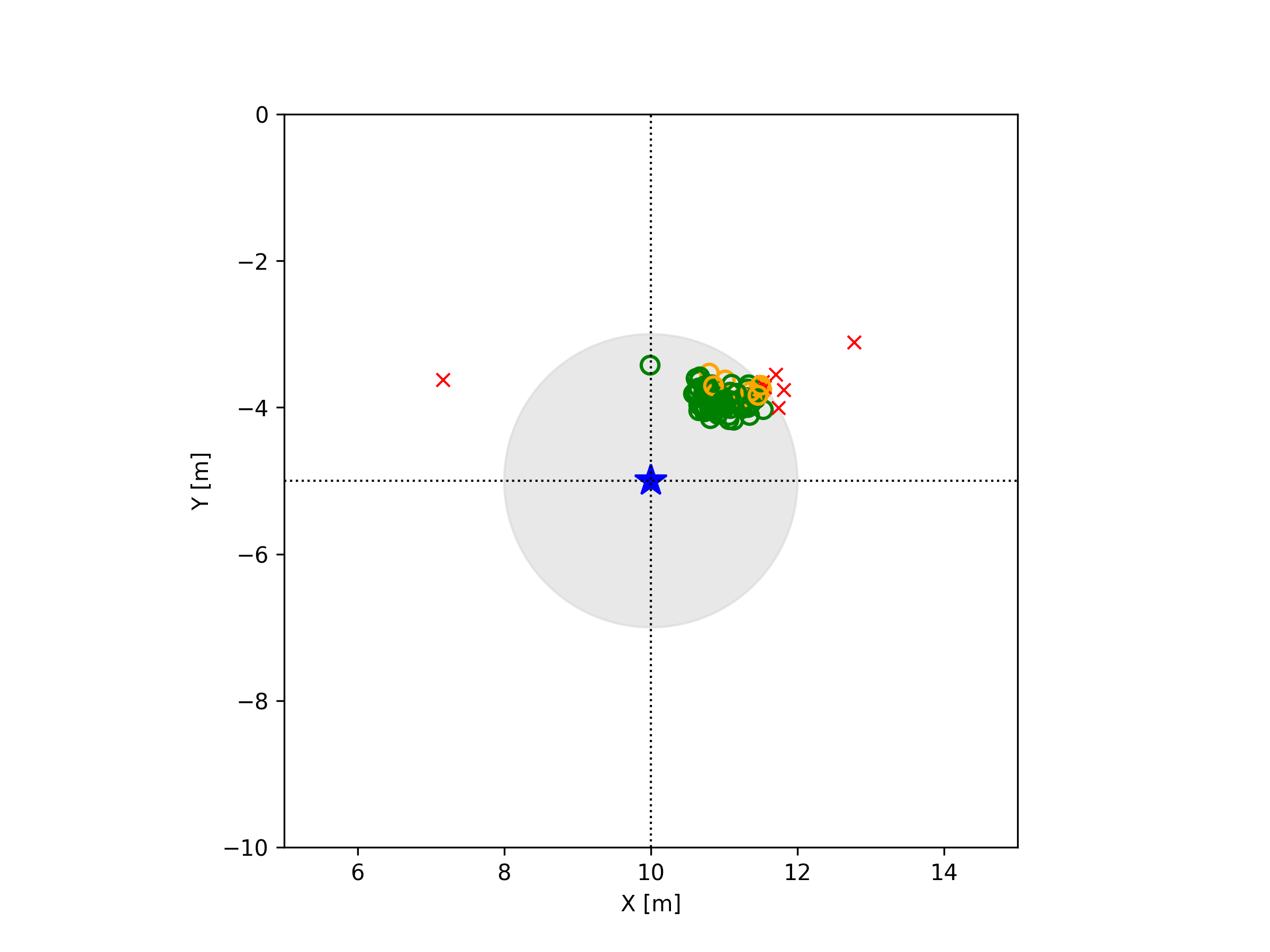}
        \caption{Landing points for the 100 simulations with LZ at $[10,-5]$.}
       \label{fig:LZ_10-5}
    \end{subfigure}
    \caption{Top-down view of the LZ with the designated landing points of each of the evaluations simulations, showing whether it was a successful, partially successful, or failed landing.}
    \label{fig:LZ}
 \end{figure}

For the second set of trajectories, 81 successfully landed, 12 had an increased landing norm velocity between $0.01$ and $1.24\,\mathrm{ms^{-1}}$ larger than the threshold, whilst 7 crashed completely. Of the 12 that missed the landing velocity, the largest $z$ velocity was only $0.91\,\mathrm{ms^{-1}}$ greater than the threshold, hence the primary cause of failure was due to an excessive $x$ direction velocity.  Figures \ref{fig:LZ_0} and \ref{fig:LZ_10-5} depict each of these scenarios in the evaluation simulations to emphasize the precise nature of DreamLander. Successful landing is marked by a green circle, missing due to position is marked by a yellow circle, crashing due to velocity is marked by an orange circle and a complete miss and crash is marked by a red cross. 

The increased number of failure cases in the second set of trajectories can be attributed to uneven terrain, causing premature collision because of the inherent policy that is learned, since DreamLander appears to hover close to the ground and then translate toward the LZ. Although undesirable, through tuning of the reward function in future work, this can be alleviated.

\begin{figure}[H]
    \centering
    \begin{subfigure}[t]{0.3\textwidth}
        \centering
        \includegraphics[width=\linewidth]{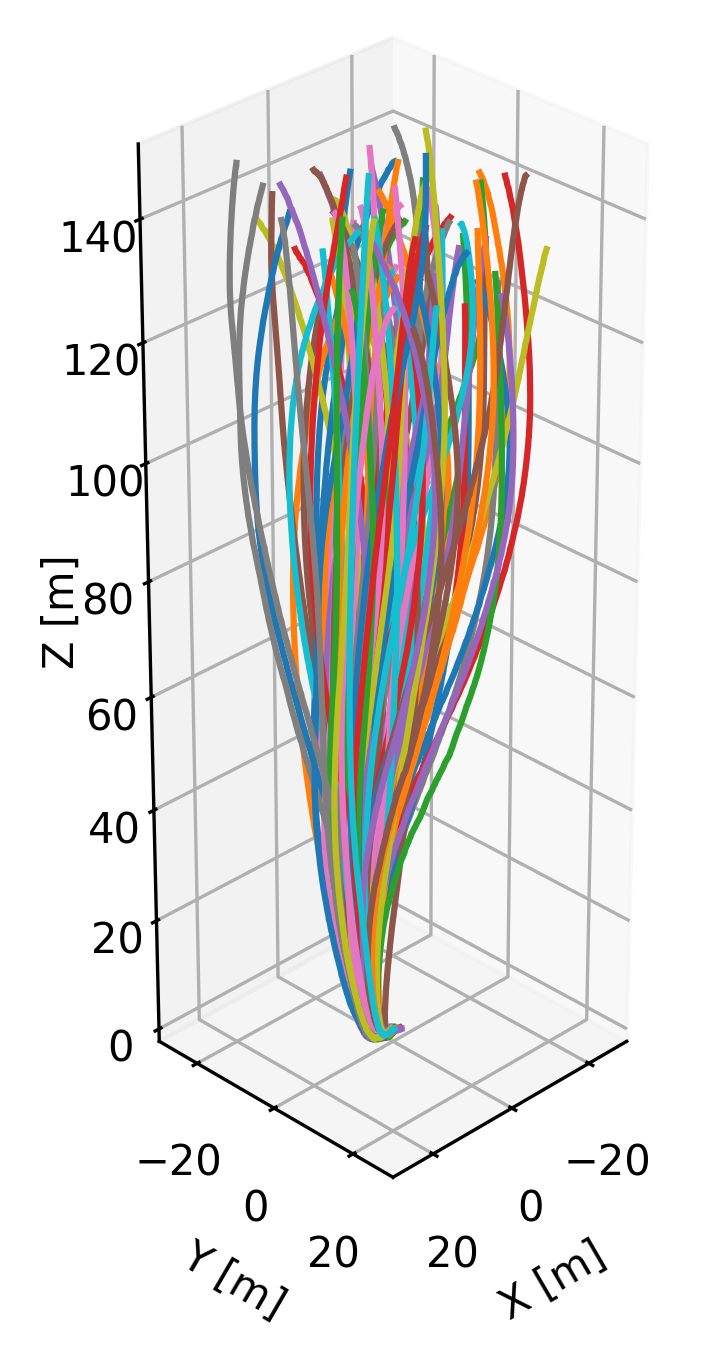}
        \caption{3D view of each of the evaluation trajectories with the origin as the designated LZ.}
       \label{fig:3dtraj_0}
    \end{subfigure}
    \hspace{65pt}
    \begin{subfigure}[t]{0.3\textwidth}
        \centering
        \includegraphics[width=\linewidth]{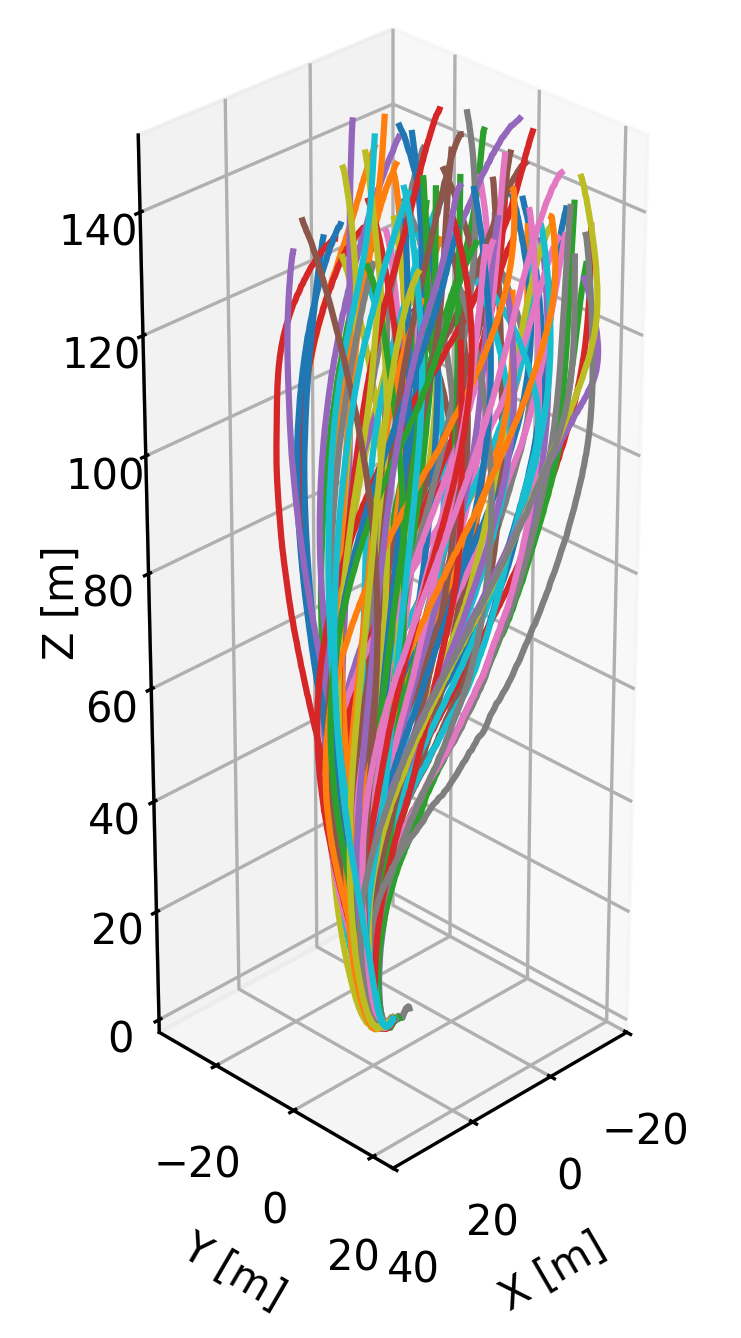}
        \caption{3D view of each of the evaluation trajectories with $[10,-5]$ as the designated LZ.}
       \label{fig:3dtraj_10-5}
    \end{subfigure}
    \caption{3D view of the 200 total evaluation trajectories.}
    \label{fig:LZ}
 \end{figure}

Figures \ref{fig:3dtraj_0} and \ref{fig:3dtraj_10-5} show the landing trajectories of the 200 evaluation simulations and that every trajectory wass able to regulate toward the LZ as would be expected. We note the set of evaluation trajectories where the coordinates of the LZ were set to the location $[10,-5]$, to highlight the ability of the model to regulate not only to the origin but also to any desired location, which will be important when the LZ is dynamically allocated based on pixel observations in future work. 

\begin{figure*}[htbp]
    \centering

    \newcolumntype{C}{>{\centering\arraybackslash}m{0.3\linewidth}} 

    \begin{tabular}{C C C}

        \includegraphics[width=\linewidth]{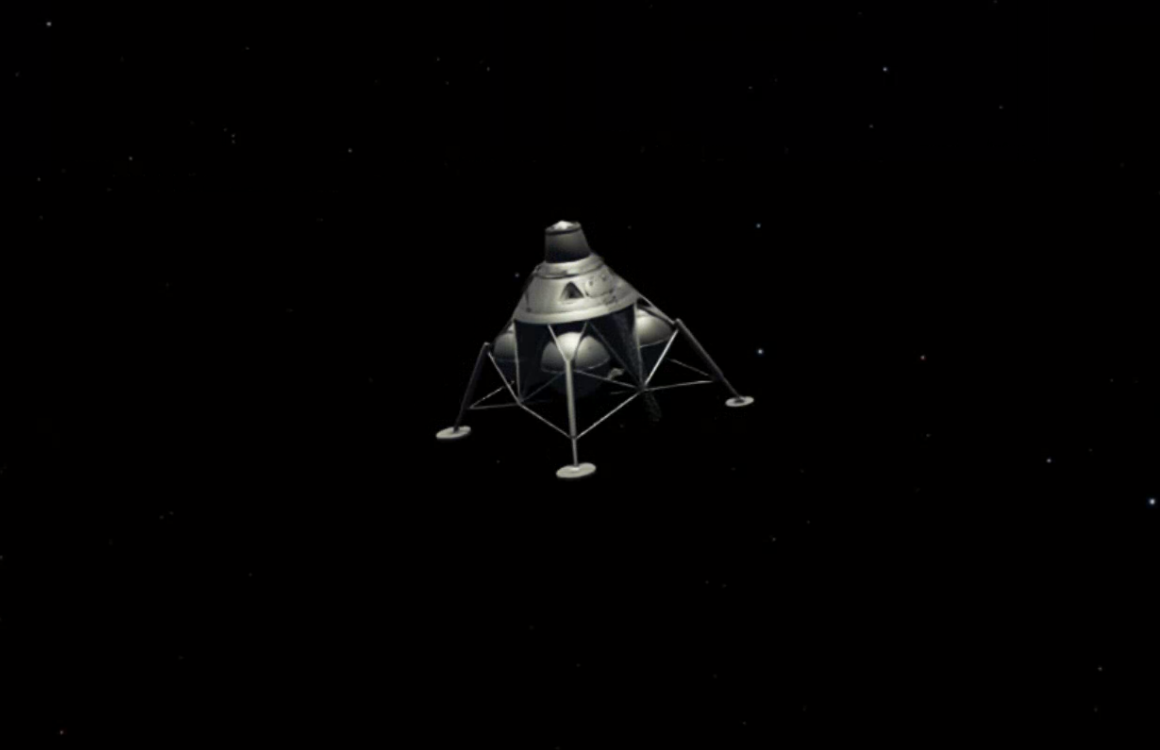} &
        \includegraphics[width=\linewidth]{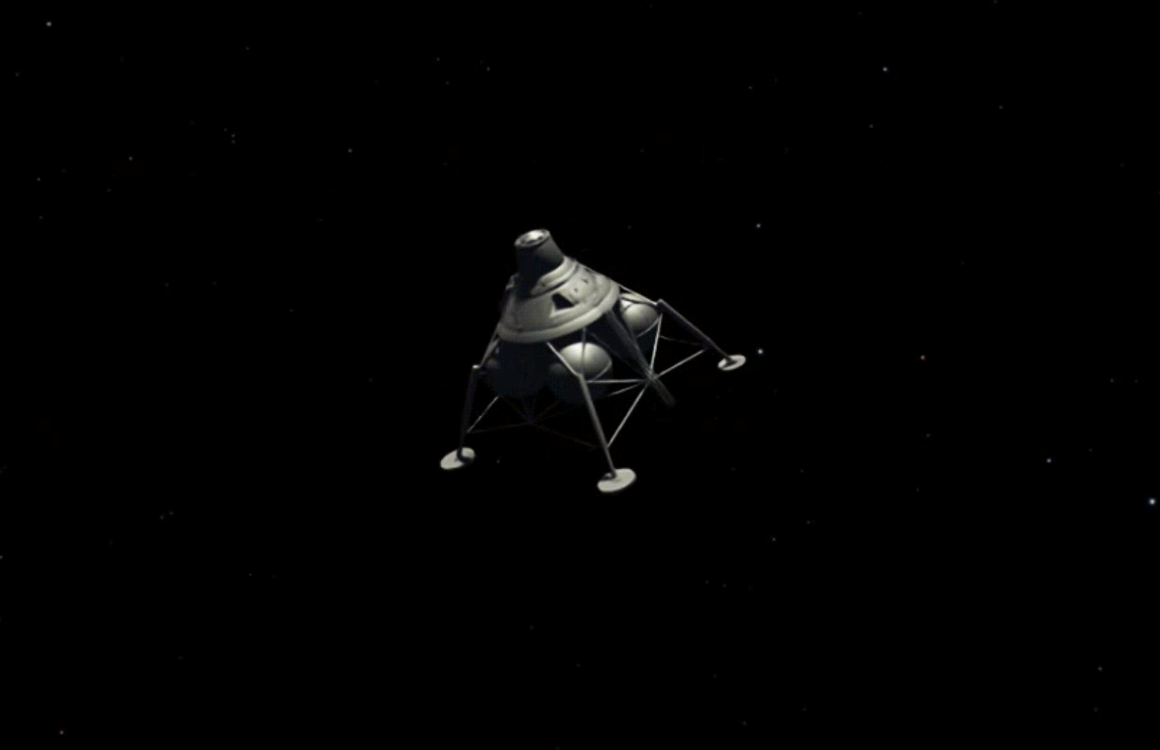} &
        \includegraphics[width=\linewidth]{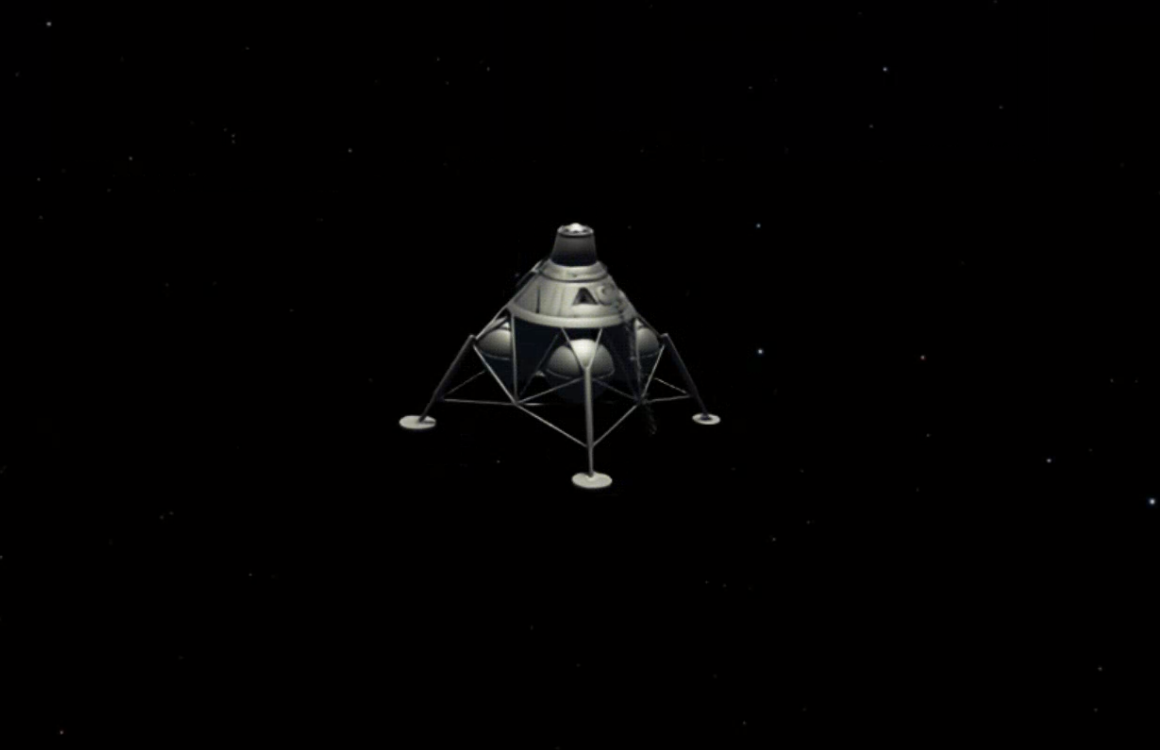} \\
        \addlinespace[1pt] 
        \textbf{$t_1$} & \textbf{$t_2$} & \textbf{$t_3$} \\
        \addlinespace[4pt] 
        \includegraphics[width=\linewidth]{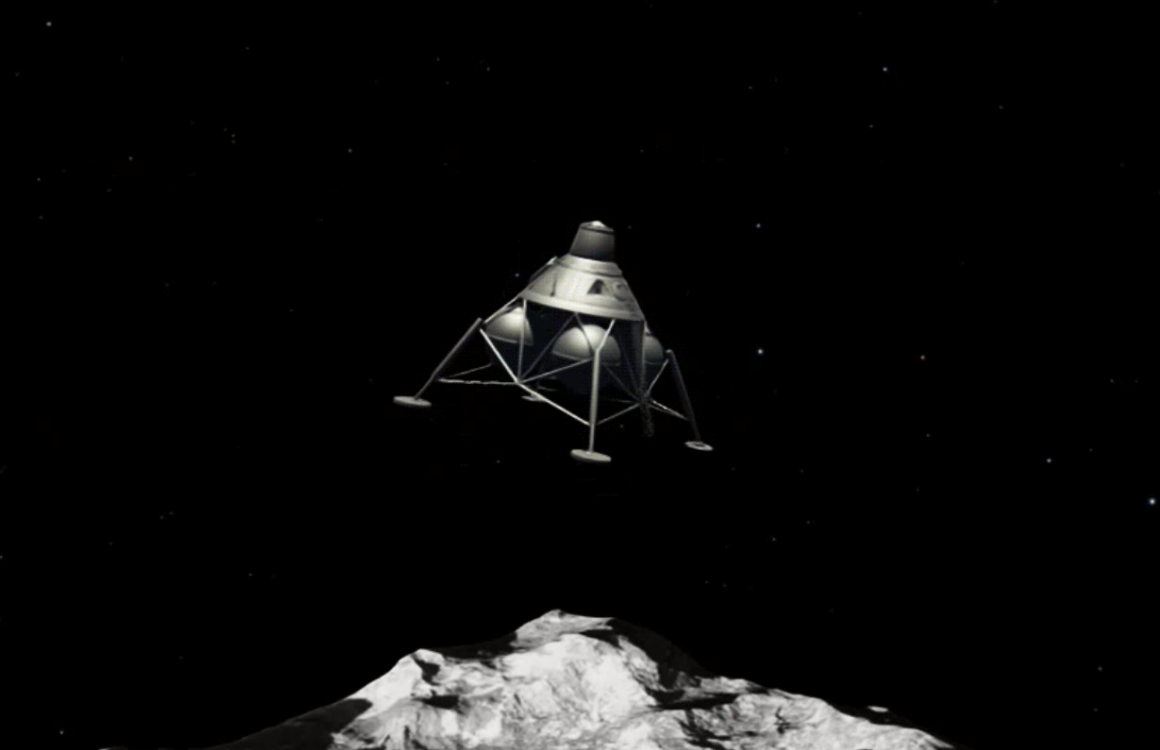} &
        \includegraphics[width=\linewidth]{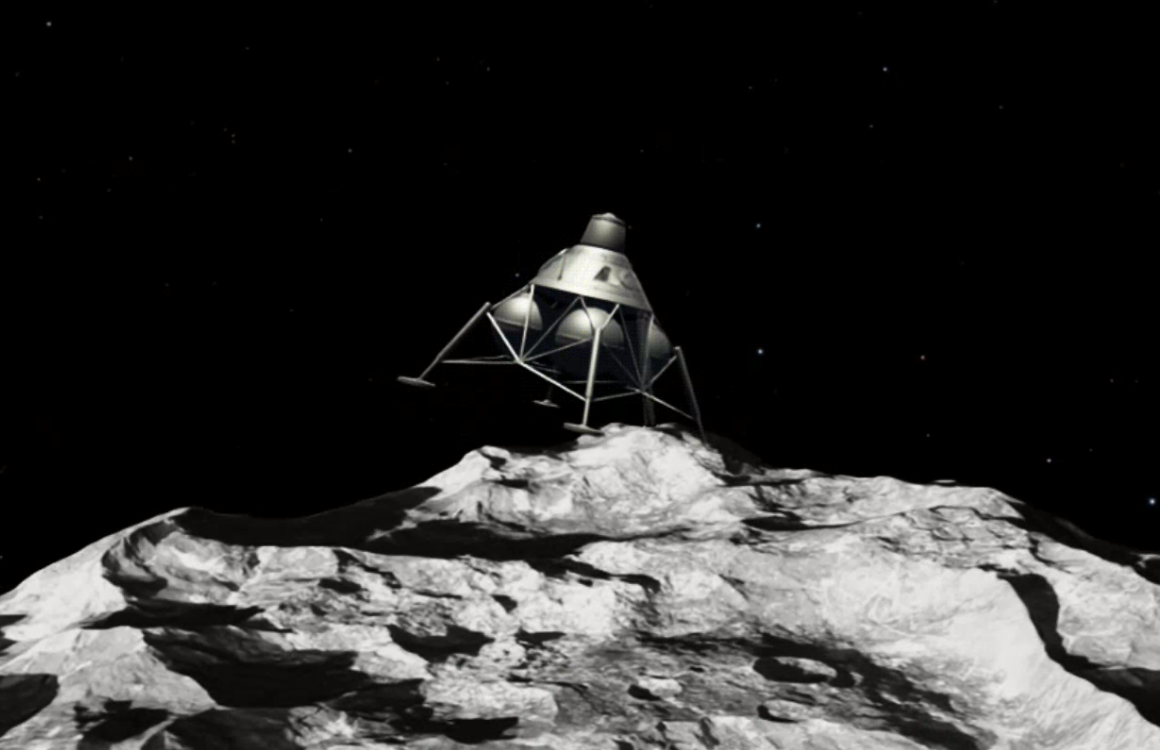} &
        \includegraphics[width=\linewidth]{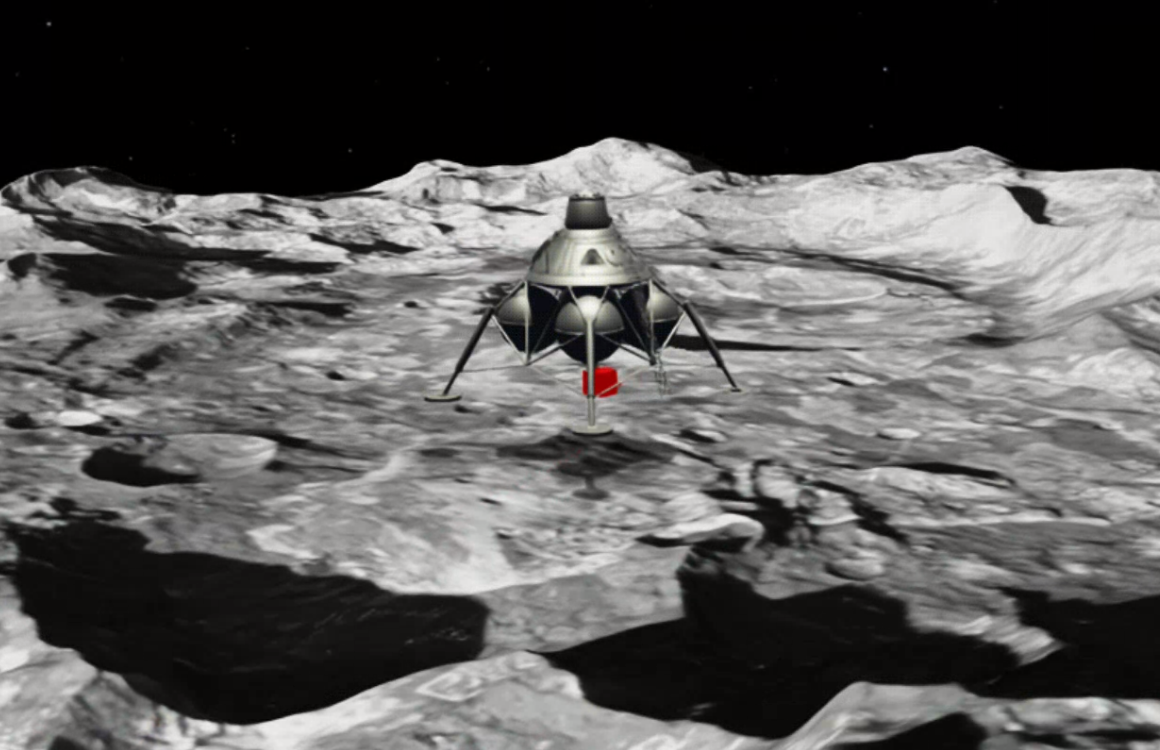} \\
        \addlinespace[1pt] 
        \textbf{$t_4$} & \textbf{$t_5$} & \textbf{$t_6$}\\
        
        \addlinespace[4pt] 

    \end{tabular}
    \caption{Snapshots of a landing trajectory of DreamLander, highlighting its use of attitude control to utlize the main engine thrust in positional alignment.}
    \label{fig:ThrustAlign}
\end{figure*}

A particularly interesting control strategy emerges as a result of DreamLander's training. Whilst the initial position in the $x$ and $y$ directions is not substantially large, it can be large enough that simply using the force from the RCS thrusters may not allow the vehicle to converge precisely to the LZ. As such, whilst the vehicle is decreasing in altitude, full main engine use is not explicitly needed, DreamLander reorients the lander to tilt toward the direction of the LZ, and fire the main engine to allow for faster directional alignment, whilst also maintaining an appropriate descent rate. This effect can be seen in the both rows of images in Figure \ref{fig:ThrustAlign}. This control strategy is both intuitive and akin to thrust vectoring controlled landing that is utilized by SpaceX in the recapture of their boosters \cite{6dofrocket} and other lander vehicles \cite{aerospace10070590}. While we impose a penalty on the alignment of the vehicle with the ENU axis, balancing the rewards to achieve a plausible yet optimal control strategy is key. 

\begin{figure}[h!]
    \centering
    \includegraphics[width=0.9\linewidth]{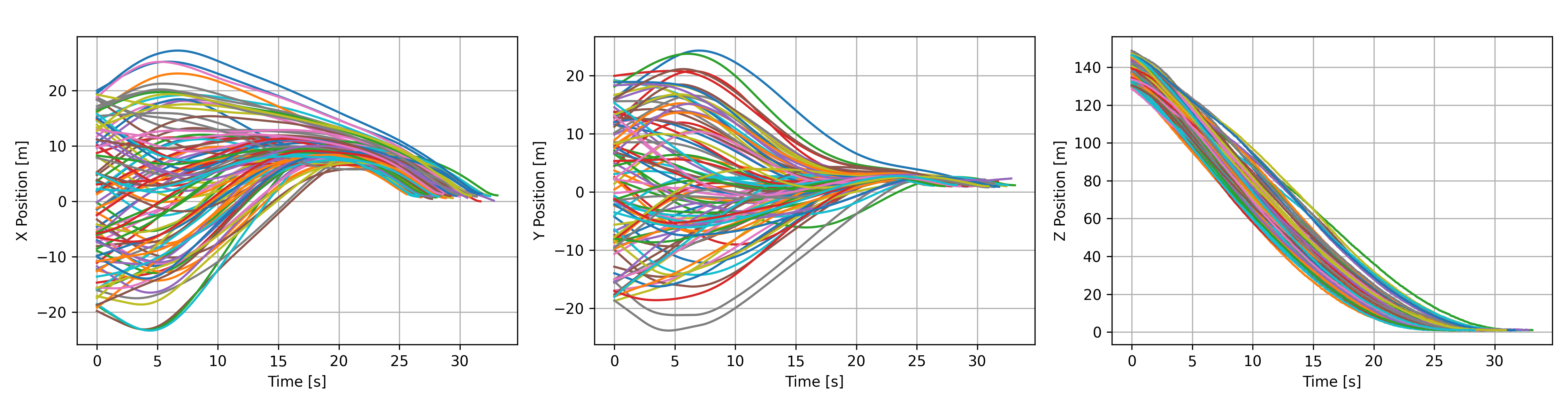}
    \caption{Position trajectories in the x,y and z directions of each of the evaluation simulations.}
    \label{fig:position}
\end{figure}

\begin{figure}[h!]
    \centering
    \includegraphics[width=0.9\linewidth]{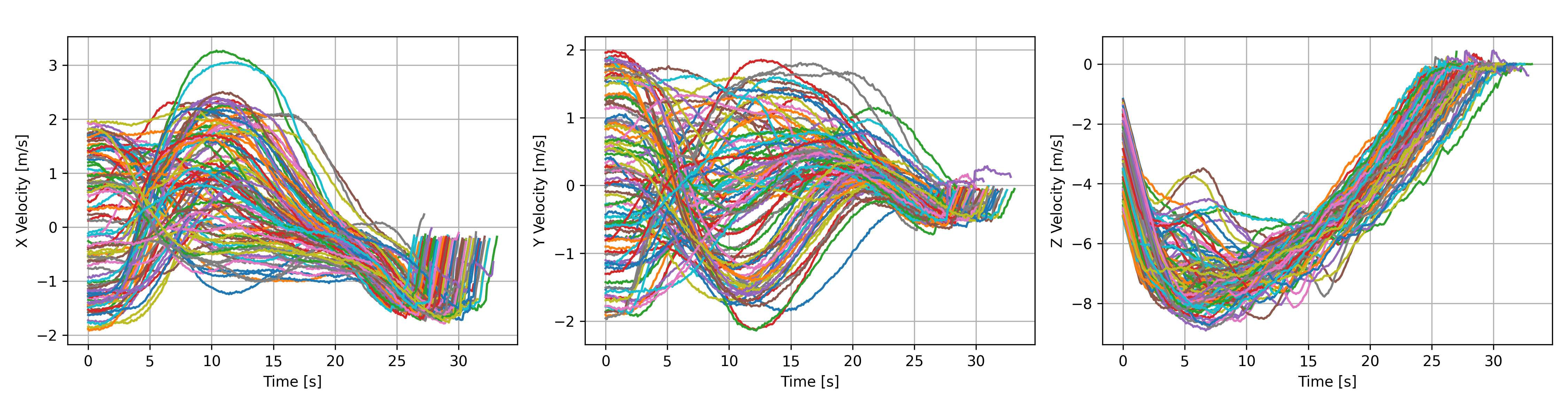}
    \caption{Velocity trajectories in the x,y and z directions of each of the evaluation simulations.}
    \label{fig:velocity}
\end{figure}

\begin{figure}[h!]
    \centering
    \includegraphics[width=0.9\linewidth]{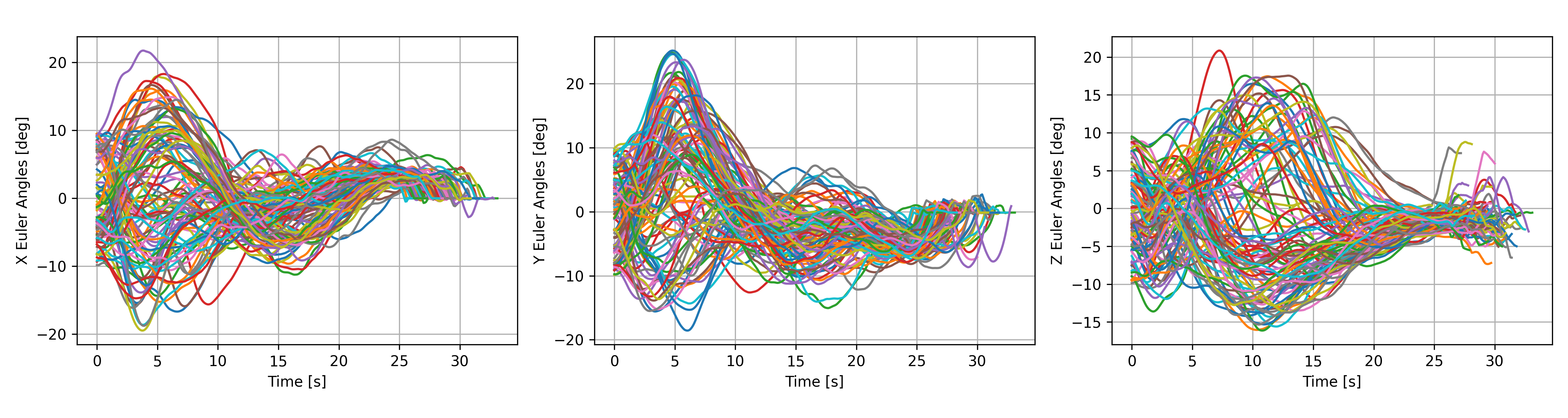}
    \caption{Euler angle trajectories in the roll, pitch, yaw directions of each of the evaluation simulations.}
    \label{fig:eulerangles}
\end{figure}

In order to demonstrate the capability of DreamLander to generate trajectories that are smooth and trackable by when used as a guidance algorithm, we display the following position and velocity trajectories. Figure \ref{fig:position} displays the position trajectories for all three axes for each of the evaluation simulations, while Figure \ref{fig:velocity} displays the velocities. Although we penalize DreamLander for not aligning with the ENU frame, because of the natural control strategy it learns, throughout the trajectory, the vehicle may need to deviate from exact alignment, hence why the trajectory of the Euler angles does not exponentially converge like the position and velocities. Figure \ref{fig:eulerangles} shows this and the fact that despite this fact, they each stay close to zero.

\pagebreak

\begin{figure*}[h!]
    \centering

    \newcolumntype{C}{>{\centering\arraybackslash}m{0.3\linewidth}} 

    \begin{tabular}{C C C}
        
        \text{Dynamics Predictor} & \text{State Predictor} & \text{Reward Predictor} \\
        \includegraphics[width=\linewidth]{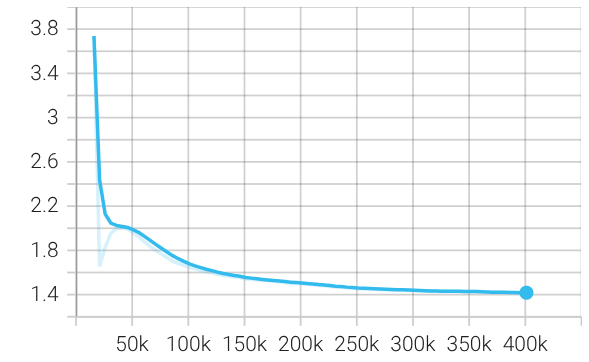} & 
        \includegraphics[width=\linewidth]{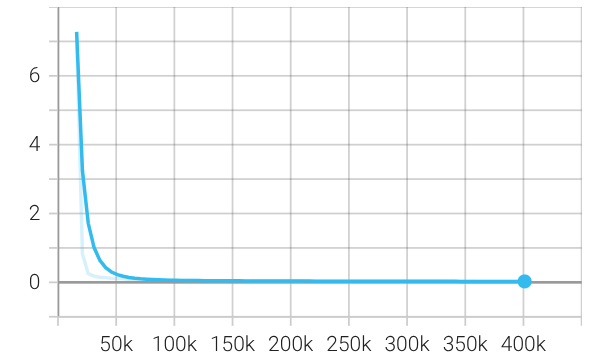} &
        \includegraphics[width=\linewidth]{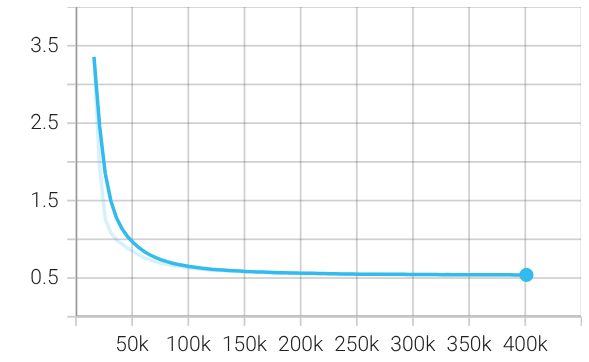} \\
        \text{Iterations} & \text{Iterations} & \text{Iterations} \\
        \addlinespace[1pt] 
        \text{Actor Loss} & \text{Value Loss} & \text{Continuation Predictor} \\
        \includegraphics[width=\linewidth]{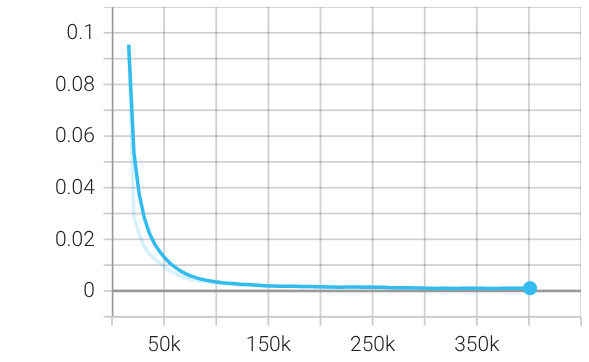} &
        \includegraphics[width=\linewidth]{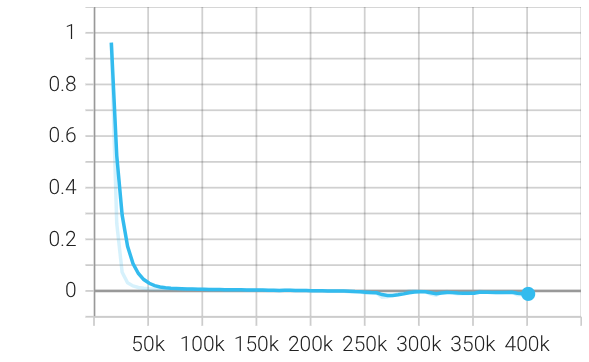} & 
        \includegraphics[width=\linewidth]{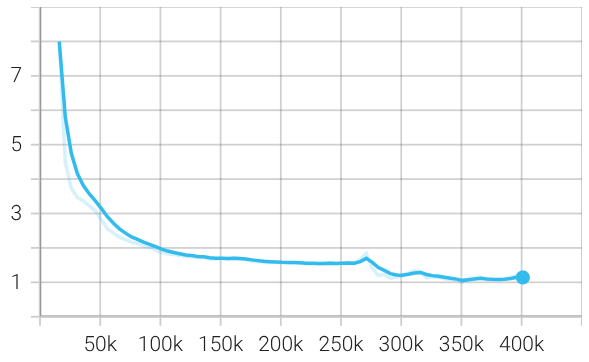}\\
        \text{Iterations} & \text{Iterations} & \text{Iterations} \\
        \addlinespace[1pt] 
        \addlinespace[4pt] 

    \end{tabular}
    \caption{Training Loss plots for each of the key predictors, actor and value functions for DreamLander, which is based on the DreamerV3 architecture.}
    \label{fig:training_plots}
\end{figure*}

In order to demonstrate the convergence of the training losses of both the models and prediction heads of DreamLander, we include the following plots, left to right, top to bottom in Figure \ref{fig:training_plots}: dynamic prediction loss, state prediction loss, reward prediction loss, continuation prediction loss, actor loss and value loss. 

\section{Conclusion}

In this work we present DreamLander, a DreamerV3 based MBRL framework capable of a highly precise and efficient planetary landing control policy. It has shown to capably learn the dynamics of the EDL problem and then formulate a control policy capable of landing the vehicle with pinpoint accuracy. The model trains efficiently on data provided in simulation and with the latent imagination it produces. We demonstrated its capability to successfully solve the EDL over a large range of initial conditions, and learn a policy that is intuitive. The model based nature of this framework also enables the further inclusion of safety based algorithms to be included so that undesired states can be avoided, further demonstrating the capability of MBRL applications in space related domains. Future work can also investigate the ability of this architecture to learn in real-time or with few-shot simulations to generalize to other environments such as the Moon or Mars, or the inclusion of pixel based observations to improve the autonomy of the guidance algorithm.

\nocite{*}
\bibliography{main}

\end{document}